\documentclass[letterpaper, 10 pt]{IEEEtran}
\usepackage{etex}
\IEEEoverridecommandlockouts  

\usepackage{multicol}
\usepackage[bookmarks=true]{hyperref}

\usepackage{comment}
\usepackage{siunitx}
\usepackage{relsize}
\usepackage{ifthen}

\usepackage[caption=false]{subfig}

\usepackage[vlined,ruled]{algorithm2e}
\usepackage{graphics} 
\usepackage{rotating}
\usepackage{color}
\usepackage{enumerate}
\usepackage[T1]{fontenc}
\usepackage{psfrag}
\usepackage{epsfig} 
\usepackage{booktabs}
\usepackage{graphicx,url}
\usepackage{multirow}
\usepackage{array}
\usepackage{latexsym}
\usepackage{amsfonts}
\usepackage{amsmath}
\usepackage{amssymb}
\usepackage{xstring}
\usepackage[noend]{algorithmic}
\usepackage{multirow}
\usepackage{xcolor}
\usepackage{prettyref}
\usepackage{flexisym}
\usepackage{bigdelim}
\usepackage{breqn} 
\usepackage{listings}
\let\labelindent\relax
\usepackage{enumitem}
\usepackage{xspace}
\usepackage{bm}
\graphicspath{{./figures/}}
\usepackage{tikz}
\usetikzlibrary{matrix,calc}

\usepackage{mdwlist}

\let\itemize\undefined
\makecompactlist{itemize}{stditemize}

\newrefformat{prob}{Problem\,\ref{#1}}
\newrefformat{sec}{Section\,\ref{#1}}
\newrefformat{sub}{Section\,\ref{#1}}
\newrefformat{prop}{Proposition\,\ref{#1}}
\newrefformat{app}{Appendix\,\ref{#1}}
\newrefformat{alg}{Algorithm\,\ref{#1}}
\newrefformat{cor}{Corollary\,\ref{#1}}
\newrefformat{thm}{Theorem\,\ref{#1}}
\newrefformat{fig}{Fig.\,\ref{#1}}
\newrefformat{tab}{Table\,\ref{#1}}

\newtheorem{theorem}{Theorem}

\newtheorem{lemma}[theorem]{Lemma}

\newtheorem{proposition}[theorem]{Proposition}

\newcommand{\bdmath}{\begin{dmath}}
\newcommand{\edmath}{\end{dmath}}
\newcommand{\beq}{\begin{equation}}
\newcommand{\eeq}{\end{equation}}
\newcommand{\bdm}{\begin{displaymath}}
\newcommand{\edm}{\end{displaymath}}
\newcommand{\bea}{\begin{eqnarray}}
\newcommand{\eea}{\end{eqnarray}}
\newcommand{\beal}{\beq \begin{array}{ll}}
\newcommand{\eeal}{\end{array} \eeq}
\newcommand{\beas}{\begin{eqnarray*}}
\newcommand{\eeas}{\end{eqnarray*}}
\newcommand{\ba}{\begin{array}}
\newcommand{\ea}{\end{array}}
\newcommand{\bit}{\begin{itemize}}
\newcommand{\eit}{\end{itemize}}
\newcommand{\ben}{\begin{enumerate}}
\newcommand{\een}{\end{enumerate}}

\newcommand{\calC}{{\cal C}}

\newcommand{\calE}{{\cal E}}

\newcommand{\calG}{{\cal G}}

\newcommand{\calL}{{\cal L}}

\newcommand{\calV}{{\cal V}}

\newcommand{\setal}{~\emph{et~al.}\xspace}

\newcommand{\M}[1]{{\bm #1}} 
\renewcommand{\boldsymbol}[1]{{\bm #1}}

\newcommand{\hide}[1]{}

\newcommand{\hiddenText}{{\color{gray} hidden text.}}
\newcommand{\hideWithText}[1]{\hiddenText}

\newcommand{\subject}{\text{subject to}}

\newcommand{\normsq}[2]{\left\|#1\right\|^2_{#2}}

\newcommand{\prob}[1]{{\mathbb P}\left(#1\right)}
\newcommand{\tran}{^{\mathsf{T}}}

\newcommand{\trace}[1]{\mathrm{tr}\left(#1\right)}

\newcommand{\e}{{\mathrm e}}

\newcommand{\zero}{{\mathbf 0}}
\newcommand{\eye}{{\mathbf I}}
\newcommand{\vect}[1]{\left[\begin{array}{c}  #1  \end{array}\right]}
\newcommand{\matTwo}[1]{\left[\begin{array}{cc}  #1  \end{array}\right]}

\newcommand{\Real}[1]{ { {\mathbb R}^{#1} } }

\newcommand{\at}[1]{^{(#1)}}

\newcommand{\setdef}[2]{ \{#1 \; {:} \; #2 \} }

\newcommand{\SEthree}{\ensuremath{\mathrm{SE}(3)}\xspace}

\newcommand{\SOthree}{\ensuremath{\mathrm{SO}(3)}\xspace}
\newcommand{\Othree}{\ensuremath{\mathrm{O}(3)}\xspace}
\newcommand{\SOn}{\ensuremath{\mathrm{SO}(n)}\xspace}

\newcommand{\intlogmap}[1]{\mathrm{Log}\left(#1\right)}

\newcommand{\logmap}[1]{\intlogmap{#1}}

\newcommand{\MA}{\M{A}}

\newcommand{\ME}{\M{E}}

\newcommand{\MM}{\M{M}}

\newcommand{\MQ}{\M{Q}}

\newcommand{\MR}{\M{R}}
\newcommand{\MS}{\M{S}}

\newcommand{\MH}{\M{H}}
\newcommand{\ML}{\M{L}}

\newcommand{\MT}{\M{T}}

\newcommand{\MOmega}{\M{\Omega}}

\newcommand{\vb}{\boldsymbol{b}}

\newcommand{\vr}{\boldsymbol{r}}

\newcommand{\vt}{\boldsymbol{t}}
\newcommand{\vxx}{\boldsymbol{x}}

\newcommand{\vlambda}{\boldsymbol{\lambda}}

\newcommand{\sumalledges}{
     \displaystyle
     \sum_{(i,j) \in \calE}}

\newcommand{\scenario}[1]{{\smaller \sf#1}\xspace}

\newcommand{\grid}{\scenario{cube}}
\newcommand{\rim}{\scenario{rim}}
\newcommand{\cubicle}{\scenario{cubicle}}
\newcommand{\sphere}{\scenario{sphere}}
\newcommand{\sphereHard}{\scenario{sphere-a}}
\newcommand{\garage}{\scenario{garage}}
\newcommand{\torus}{\scenario{torus}}

\newcommand{\SDPA}{\scenario{SDPA}}

\usepackage{float}

\usepackage{epstopdf}
\usepackage{epsfig}

\newcommand{\extended}[1]{}

\newcommand{\lam}{\lambda}
\newcommand{\vlam}{\vlambda}

\newcommand{\forJournal}[1]{}

\graphicspath{{./figures/}}

\newcommand{\nrNodes}{n}
\newcommand{\nrObsNodes}{n-1}

\newcommand{\MRij}{\bar{\MR}_{ij}}

\newcommand{\MRlift}{{\MQ}}
\newcommand{\MRijlift}{\MRlift_{ij}}

\newcommand{\vtij}{\bar{\vt}_{ij}}
\newcommand{\vtlift}{{\MT}}
\newcommand{\vtijlift}{\vtlift_{ij}}

\newcommand{\MAlift}{\breve{\MA}}
\newcommand{\MAanc}{\MA}

\newcommand{\MElift}{\breve{\ME}}
\newcommand{\MEanc}{{\ME}}

\newcommand{\lamPara}{\lam} 
\newcommand{\lamPerp}{\lam} 
\newcommand{\lamy}{\lam_y}

\newcommand{\lamParaStar}{\lamPara^\star}
\newcommand{\lamPerpStar}{\lamPerp^\star}
\newcommand{\lamyStar}{\lamy^\star}
\newcommand{\vlamStar}{\vlambda^{\star}}
\newcommand{\vlamHat}{\hat{\vlambda}}

\newcommand{\vxxStar}{\vxx^\star}
\newcommand{\yStar}{y^\star}

\newcommand{\vdxij}{{\bar{\vxx}}_{ij}}
\newcommand{\frob}{{\tt F}}

\newcommand{\fML}{f_{\tt ML}}
\newcommand{\fMLstar}{f_{\tt ML}^\star}

\newcommand{\PGO}{PGO\xspace}

\newcommand{\normal}{\mbox{gaussian}}
\newcommand{\vonmises}{\mbox{vonMises}}

\renewcommand{\ML}{ML\xspace}
\renewcommand{\gamma}{\lambda^\circ}
\newcommand{\rows}{\mbox{rows}}
\newcommand{\candidate}{\hat{\vxx}}

\newcommand{\obsNodes}{1,\ldots,\nrNodes-1}
\newcommand{\sumObsNodes}{\sum_{i=1}^{\nrNodes-1}}

\newcommand{\verificationOne}{\scenario{V1\xspace}}
\newcommand{\verificationTwo}{\scenario{V2\xspace}}

\newcommand{\sigmaT}{\sigma_T}
\newcommand{\sigmaR}{\sigma_R}
\newcommand{\fopt}{f^\star}
\newcommand{\fsubopt}{f^\dag}
\newcommand{\xopt}{\vxx^\star}
\newcommand{\xsubopt}{\vxx^\dag}

\newcommand{\journalVersion}[1]{}

\usepackage{xr}
\title{\huge{Lagrangian Duality in 3D SLAM: \\ Verification Techniques and Optimal Solutions}}
 \author{Luca Carlone, \; David M. Rosen, \; Giuseppe Calafiore, \; John J. Leonard, \; Frank Dellaert
 \thanks{
 L.\,Carlone and F.\,Dellaert are with the Georgia Institute of Technology, Atlanta, GA, USA, 
 {\sf luca.carlone@gatech.edu}, {\sf frank@cc.gatech.edu}.
 }
 \thanks{
 D.M.\,Rosen and J.J.\,Leonard are with the Massachusetts Institute of Technology, Cambridge,
 MA, USA {\sf\{dmrosen,jleonard\}@mit.edu}. 
 }
  \thanks{
 G. Calafiore is with the Politecnico di Torino, Torino, Italy, {\sf giuseppe.calafiore@polito.it}.
 }
 }

\begin{document}

\maketitle

{\color{red} Please cite this paper as: ``L. Carlone, D.M. Rosen, G.C. Calafiore, J.J. Leonard, F. Dellaert, 
\emph{Lagrangian Duality in 3D SLAM: Verification Techniques and Optimal Solutions}, 
Int. Conf. on Intelligent RObots and Systems (IROS), 2015.''}

\vspace{0.1cm}

\begin{abstract}
State-of-the-art techniques for simultaneous localization and mapping (SLAM) employ
iterative nonlinear optimization methods to
 compute an estimate for robot poses. 
While these techniques often work well in practice, they do not provide guarantees on the 
quality of the estimate.
This paper shows that \emph{Lagrangian duality} is a powerful tool to assess the 
quality of a given candidate solution. 
Our contribution is threefold. First, we discuss a revised formulation of the SLAM inference problem. We show that this formulation is probabilistically grounded  
and has the advantage of leading to an optimization problem with quadratic 
objective.
The second contribution is the derivation of the corresponding \emph{Lagrangian dual problem}. 
The SLAM dual problem is a (convex) \emph{semidefinite program}, which can be solved reliably 
and globally by off-the-shelf solvers.  
The third contribution is to discuss the relation between the original SLAM problem and its dual.
We show that from the dual problem, one can evaluate the quality 
(i.e., the suboptimality gap) of a candidate SLAM solution, and ultimately provide 
a certificate of optimality. 
Moreover, when the duality gap is zero, one can compute a guaranteed optimal SLAM solution from the dual problem, 
circumventing non-convex optimization.
We present extensive (real and simulated) experiments supporting our claims 
and discuss practical relevance and open problems.
\end{abstract}


\section{Introduction}

\emph{Simultaneous localization and mapping} (SLAM) is an enabling technology for 
many applications, including service and industrial robotics, 
autonomous driving, search and rescue, planetary exploration,
and augmented reality. 

The last decade has witnessed several groundbreaking results in SLAM, and 
state-of-the-art approaches are now transitioning from academic research to industrial applications.
Standard techniques compute an estimate (e.g., for robot poses) 
by minimizing a nonlinear cost function, 
whose global minimum is the \emph{maximum likelihood estimate} (or \emph{maximum a posteriori} 
estimate in presence of priors). 
The optimization problem underlying SLAM is commonly solved using iterative 
nonlinear optimization methods, e.g.\ the Gauss-Newton method~\cite{Lu97,Kuemmerle11icra,Kaess12ijrr}, 
the gradient method~\cite{Olson06icra,Grisetti09its}, 
trust region methods~\cite{Rosen14tro}, or ad-hoc 
approximations~\cite{Dellaert02icra,Carlone14ijrr}.

Despite the success of state-of-the-art techniques, some practical and theoretical 
problems remain open.
While iterative approaches are observed to work well in many problem instances, 
they cannot guarantee the correctness (global optimality) of the estimates that they compute. This is due to the fact that the optimization problem is 
non-convex, hence iterative techniques may be trapped in local minima\footnote{We use the term ``local minimum'' to denote 
a stationary point of the cost which does not attain the optimal objective.} (Fig.~\ref{fig:maps}), which 
correspond to wrong estimates.  
Recent work~\cite{Carlone15icra-verify,Rosen15icra} shows that iterative 
techniques fail to converge to a correct estimate even in fairly simple 
 (real and simulated) 3D problems. Recent research efforts have
addressed the issue of global convergence from several angles.  
Olson\setal~\cite{Olson06icra}, Grisetti\setal~\cite{Grisetti09its}, Rosen\setal~\cite{Rosen14tro}, 
and Tron\setal~\cite{Tron12cdc} study iterative techniques with larger basins of convergence. 
Carlone\setal~\cite{Carlone14ijrr,Carlone14tro}, and Rosen\setal~\cite{Rosen15icra} 
propose initialization techniques to bootstrap iterative optimization.  
Huang\setal~\cite{Huang10iros}, Wang\setal~\cite{Wang12rss}, Carlone~\cite{Carlone13icra}, 
and Khosoussi\setal~\cite{Khosoussi14iros} investigate the factors influencing 
local convergence and the quality of the optimal SLAM solution. 
While these techniques provide remarkable insights into the problem, and 
working solutions to improve convergence, none of them can guarantee 
the recovery of a globally optimal solution to SLAM.


\definecolor{dgreen}{rgb}{0,0.5,0}

\newcommand{\widthCol}{3cm}

\begin{figure}[t]
\vspace{-0.5cm}
\begin{minipage}{8.5cm}
\begin{tabular}{ccc}%
& {\sphereHard} & \hspace{0.5cm} {\torus} 
\vspace{-0.1cm}
\\
\begin{sideways}{$\!\!\!\!\!\!\!\!\!\!\!$\color{blue}{Optimum}}\end{sideways} & 
\begin{minipage}{\widthCol}%
\centering%
\includegraphics[scale=0.27, trim=5cm 3cm 0cm 3cm,clip]{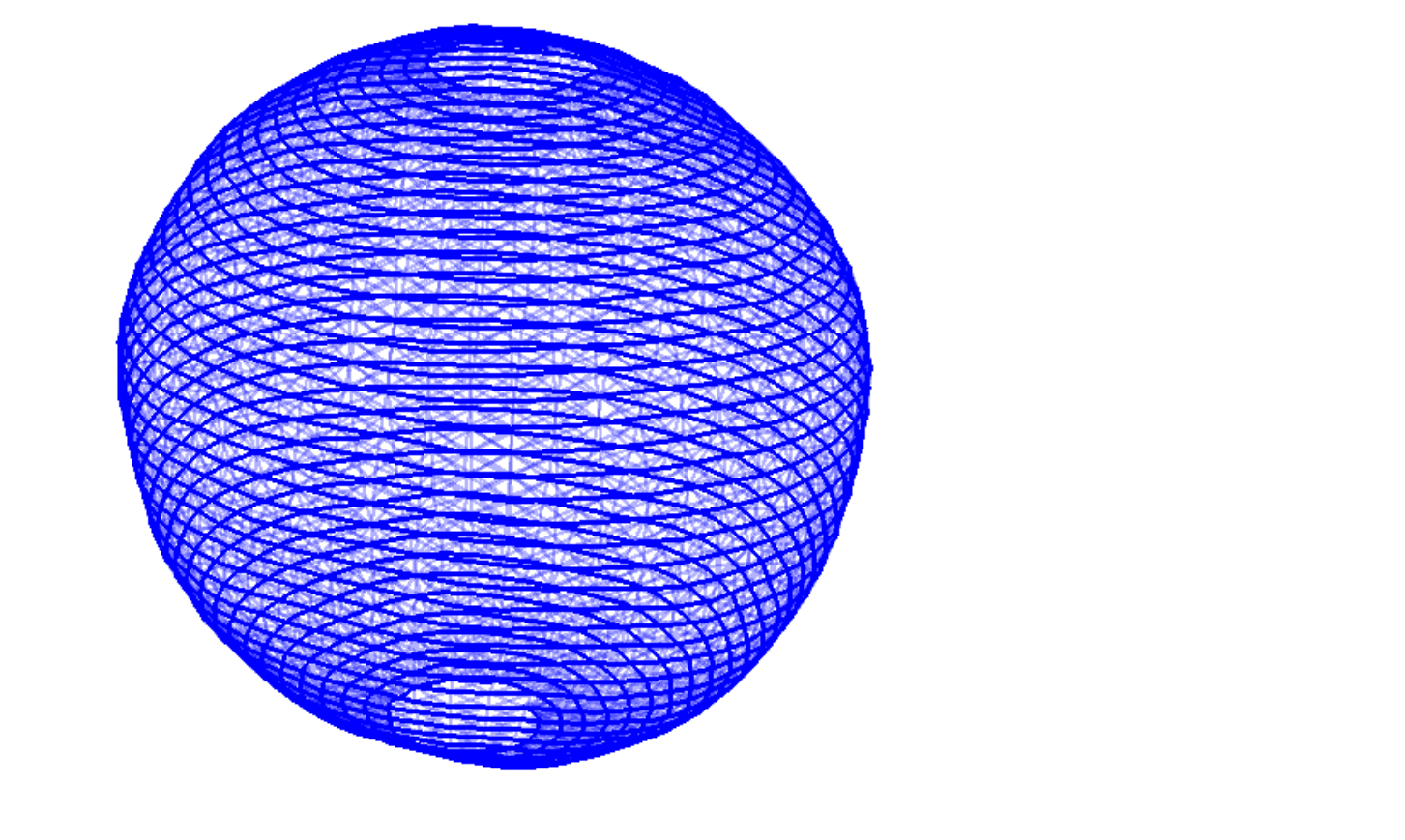} 
\end{minipage}
& 
\begin{minipage}{\widthCol}%
\centering%
\includegraphics[scale=0.3, trim=4cm 2cm 0cm 3cm,clip]{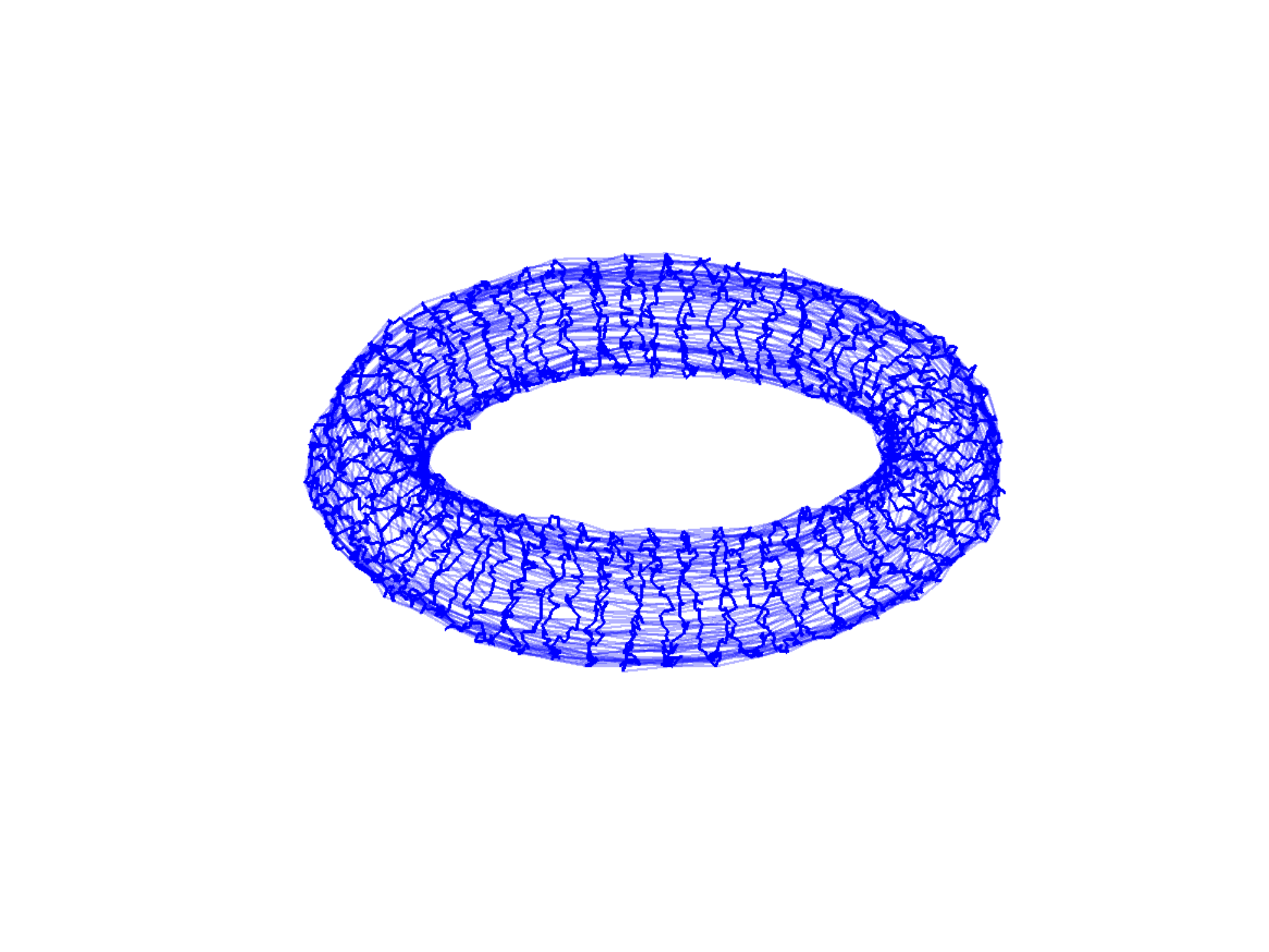}  
\end{minipage}
\vspace{-0.6cm}
\\
\begin{sideways}{$\!\!\!\!\!\!\!\!\!\!\!\!\!\!\!\!\!\!\!$\color{red}{Local Minimum}}\end{sideways} &
\begin{minipage}{\widthCol}%
\centering%
\includegraphics[scale=0.24, trim=3cm 2cm 0cm 2cm,clip]{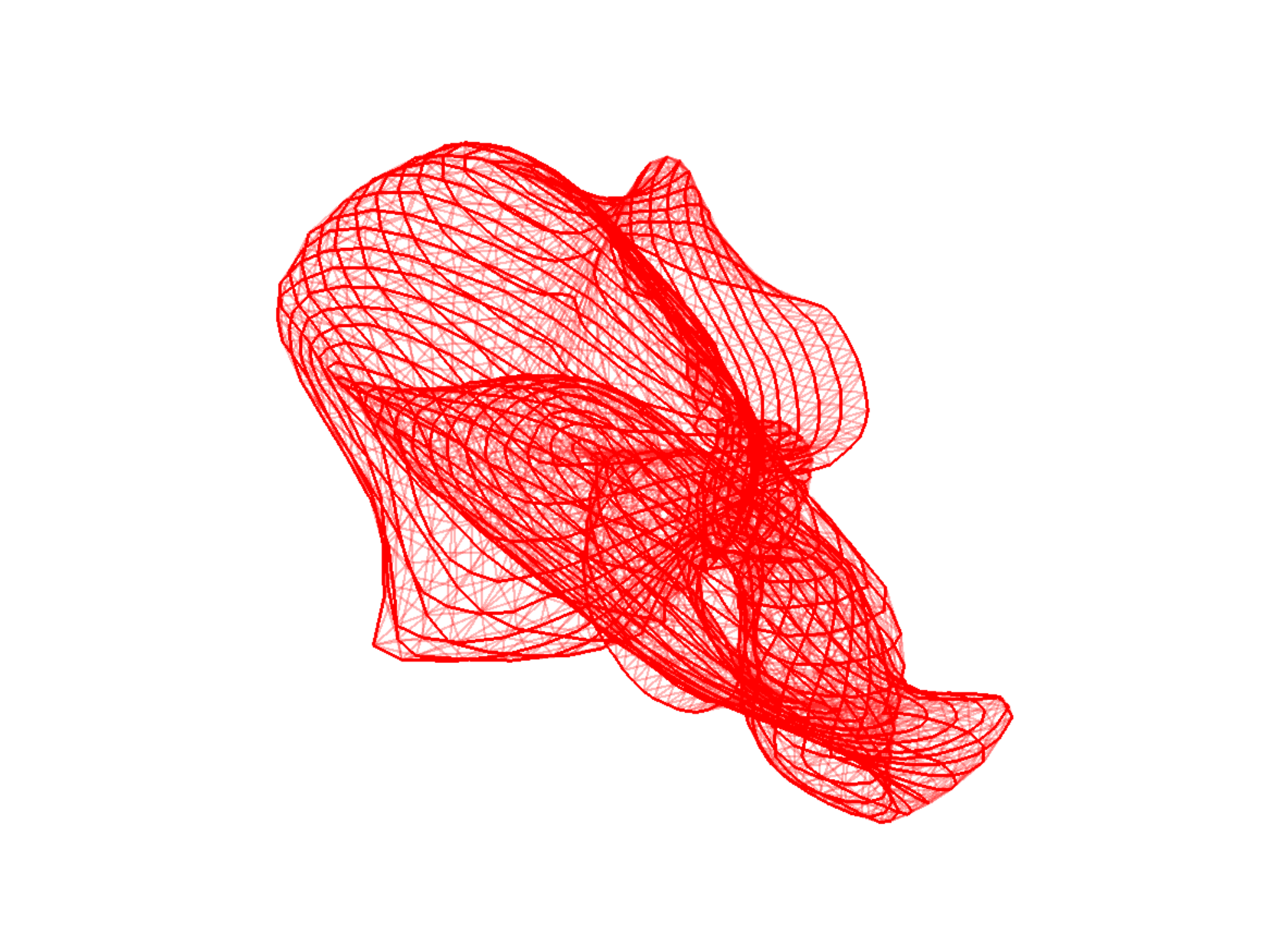} 
\end{minipage}
& 
\begin{minipage}{\widthCol}%
\centering%
\includegraphics[scale=0.32, trim=4.5cm 2cm 2cm 3cm,clip]{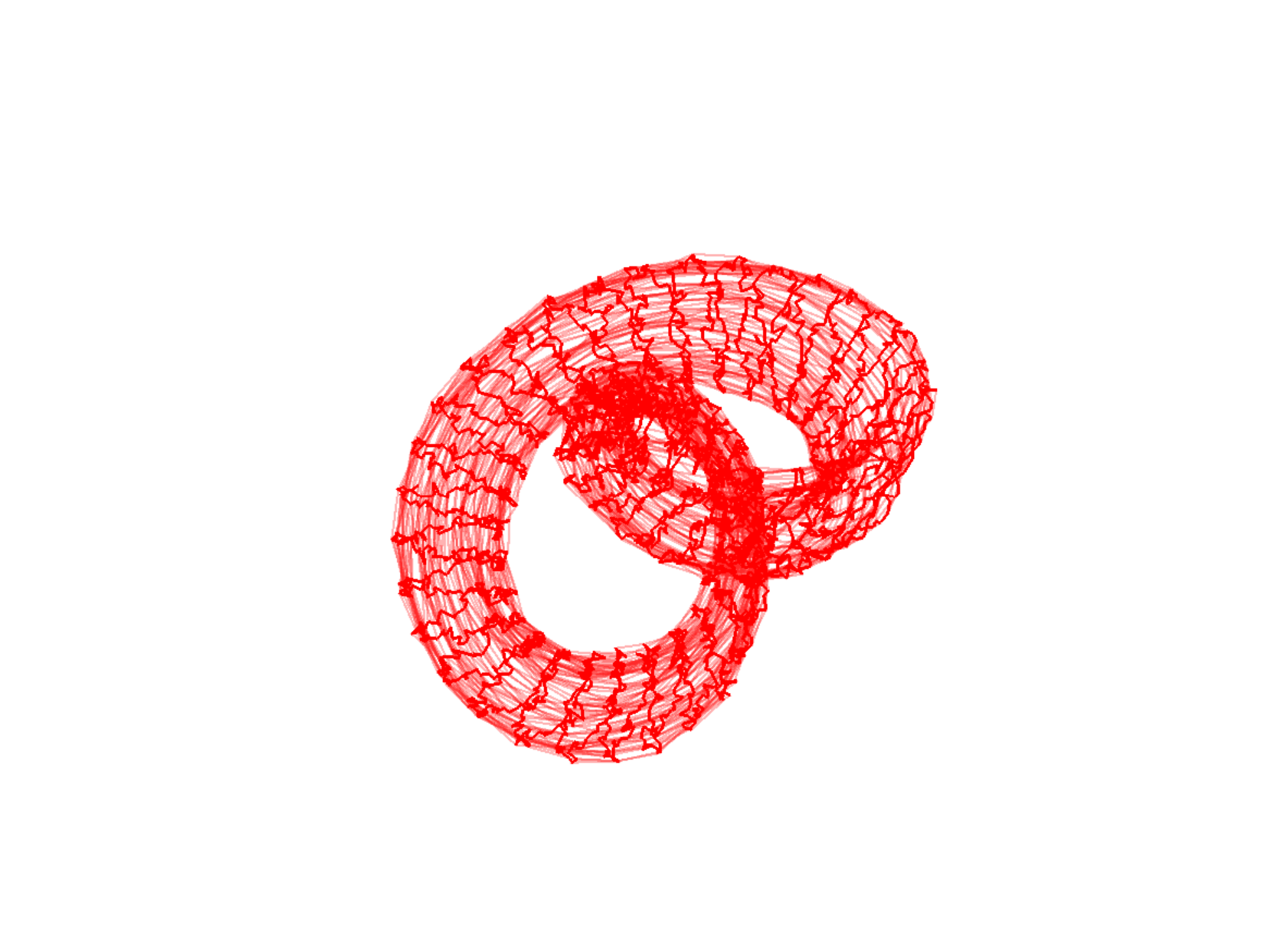}
\end{minipage}
\vspace{0.1cm}
\end{tabular}%
\vspace{-0.3cm} 
\caption{\label{fig:maps} 
This paper exploits Lagrangian duality to evaluate the 
quality of a SLAM solution. For instance, our verification techniques
are able to certify the optimality of the solutions in the first row of the figure, 
while they reject as suboptimal the estimates shown in the bottom row.
\vspace{-0.6cm} 
}
\end{minipage}%
\end{figure}

%

The motivation behind this work is that the transition of SLAM from research topic 
to industrial technology requires techniques with \emph{guaranteed performance}. 
In autonomous vehicle applications, failure to produce a correct SLAM solution 
may put passengers' lives at risk. In other applications, SLAM failures can possibly cascade 
into path planning failures (if the plan is computed using a wrong map), and 
this may prevent the reliable operation of mobile robots. 

Therefore, in this paper we address the following question: 

\noindent
{\bf Verification Problem:} 
\emph{Given a 
candidate SLAM estimate (e.g., returned by a state-of-the-art iterative solver), 
is it possible to evaluate the quality of this estimate (e.g., its sub-optimality gap), possibly 
certifying its optimality?}

To address this problem, we introduce a powerful tool, \emph{Lagrangian duality}, 
borrowing the corresponding theory from the optimization community. 
Duality was first applied to 2D SLAM by Carlone and 
Dellaert~\cite{Carlone15icra-verify}. In this work, 
we provide a nontrivial extension of~\cite{Carlone15icra-verify} to 3D SLAM. 

This paper contains three main contributions. 
The first contribution is 
 a revised formulation of the SLAM inference problem. 
 Our formulation is a probabilistically grounded maximum likelihood (\ML) estimator, and has the advantage of leading to an optimization problem with quadratic objective, 
which facilitates the derivation of the dual problem.
Our revised SLAM formulation is presented in Section~\ref{sec:revisitedPGO}.

The second contribution is the derivation of the \emph{Lagrangian dual problem}.
A key step towards this goal consists of 
rewriting the 3D SLAM problem as a quadratic optimization problem with quadratic equality constraints.
Intuitively, the constraints impose that the pose estimates are members of $\SEthree$, 
while the objective minimizes the mismatch w.r.t. the measurements.  
The dual SLAM problem, introduced in Section~\ref{sec:duality}, is a (convex) \emph{semidefinite program} (SDP), and can be solved globally by off-the-shelf solvers. 

The third contribution is to provide \emph{verification techniques} to assess 
the quality of a given SLAM solution, leveraging the relation between the standard SLAM problem and its dual.
We show that solving the dual problem allows us to bound the sub-optimality gap of a given candidate solution, 
hence we are able to quantify how far the candidate solution is from being optimal. 
Since current SDP solvers do not scale well to large problems, we also propose 
a second verification technique that does not require solving the SDP.
As a by-product of our derivation, we show that, when the duality gap is zero, 
we can compute an optimal SLAM solution directly from the dual problem.
Our verification techniques, presented in Section~\ref{sec:practicalUse},
can be seamlessly integrated in standard SLAM pipelines. Experimental evidence (Section~\ref{sec:experiments})
confirms that these techniques enable the certification of globally optimal solutions in both real and 
simulated experiments. Extra results and visualizations are given in the appendix of this paper.

\journalVersion{
While this paper, together with~\cite{Carlone15icra-verify}, constitutes a first answer to the verification problem, 
stated above, it opens other interesting research questions, that we discuss in Section~\ref{sec:conclusion}.
}



\newcommand{\omegat}{{\tiny \omega_t^2}}
\newcommand{\omegaR}{{\tiny \omega_R^2}}

\section{3D Pose Graph Optimization Revisited}
\label{sec:revisitedPGO}

We consider the \emph{pose graph optimization} (\PGO) formulation of the SLAM problem.
\PGO computes the \emph{maximum likelihood} estimate for $\nrNodes$ poses $\vxx_1,\ldots,\vxx_n$, 
given $m$ relative pose measurements $\vdxij$ between pairs of poses $i$ and $j$. 
In a 3D setup, both the unknown poses and the measurements are quantities in 
$\SEthree \doteq \{(\MR,\vt) : \MR \in \SOthree, \vt \in \Real{3 } \}$.
 We use the notation $\vxx_i\! =\! (\MR_i,\vt_i)$ and 
$\vdxij \!=\! (\MRij, \vtij)$ to make explicit the rotation
and the translation of each pose.
 \PGO can be visualized as a directed graph $\calG(\calV,\calE)$, 
in which we associate a node $i \in \calV\!=\!\{1,\!\ldots\!,n\}$ to each pose $\vxx_i$ and an edge $(i,j) \!\in\! \calE$ to each relative measurement 
$\vdxij$.

 
In this section we propose a revised \PGO formulation. 
The key difference w.r.t. related work is the use of the \emph{chordal distance} to quantify the 
rotation errors (more details in Section~\ref{sec:MLestimator}). 
 To lay the groundwork for this formulation, we begin 
 with a generative model for our measurements, and then derive the 
 corresponding \ML estimator.
 
 \subsection{Generative Noise Model}

We assume the following generative model for the relative pose 
measurements $(\MRij, \vtij)$\footnote{To keep notation simple, we consider measurements with the same distribution. 
The extension to heterogeneous 
$\omegaR$ and $\omegat$ is trivial.}:
\beq
\label{eq:measurementModel}
\hspace{-0.1cm}
\begin{array}{llllll}
\vtij \!\!\!&=&\!\!\! \MR_i\tran ( \vt_j \!-\! \vt_i  ) \!+\! \vt_{\epsilon}  &\;\; \vt_\epsilon \; \sim  \normal(\zero_3, \omegat \eye_3)  \\ 
\MRij \!\!\!&=&\!\!\! \MR_i\tran \MR_j \MR_\epsilon                           &\;\; \MR_\epsilon \sim  \vonmises(\eye_3, \omegaR)
\end{array}
\eeq
where ``$\normal(\boldsymbol{\mu}, \MOmega)$'' denotes a Gaussian distribution with mean $\boldsymbol{\mu}$ 
and \emph{information} matrix $\MOmega$, while ``$\vonmises(\MS, \kappa)$'' denotes the 
isotropic von Mises-Fisher distribution, with mean~$\MS \in \SOthree$ and~\emph{concentration parameter} 
$\kappa$. The key difference w.r.t. to measurement models in other \PGO formulations lies in the use of the 
von Mises-Fisher distribution as the model for the rotational measurements errors $\MR_\epsilon \in \SOthree$. 

The isotropic \emph{von Mises-Fisher} (or \emph{Langevin})~\cite{Boumal14ii} distribution on $\SOn$
with mean $\MS  \in \SOn$ and concentration parameter $\kappa \geq 0$
can be written explicitly as:
\beq
\label{eq:vonMises}
\prob{\MR_\epsilon} = \frac{1}{c_n(\kappa)} \exp \left( \kappa \; \trace{\MS\tran \MR_\epsilon} \right),
\eeq
where $\trace{\cdot}$ is the matrix trace and $c_n(\kappa)$ is a normalization term. Closed-form expressions for 
 $c_n(\kappa)$ 
are given in~\cite{Boumal14ii}; these are 
 inconsequential for our derivation. 
For $\kappa \rightarrow 0$, the distribution tends to the uniform distribution over $\SOn$.
For $\kappa \rightarrow \infty$, $\MR_\epsilon = \MS$ with probability one.
Roughly speaking, one may think at $\kappa$ in terms of \emph{information} content. 

We are now ready to introduce the maximum likelihood estimator for 
the poses, given the measurement model~\eqref{eq:measurementModel}.

\subsection{Maximum Likelihood Estimator}
\label{sec:MLestimator}

The \ML estimate corresponds to the set of poses maximizing the likelihood of the measurements, 
or, equivalently, minimizing the negative log-likelihood:
\beq
\label{eq:MLE}
\fMLstar = \min_{ \{ \vxx_i \in \SEthree\} } \sumalledges - \log \calL( \MRij | \vxx) - \log \calL( \vtij  | \vxx).
\eeq
%
The negative log-likelihood of the Cartesian measurements 
can be easily computed from the Gaussian distribution: 
\beq
\label{eq:nllt}
- \log \calL( \vtij  | \vxx) = \omegat \normsq{ \vt_j \!-\! \vt_i \!-\! \MR_i \vtij }{} + \mbox{const.}
\eeq
Using~\eqref{eq:vonMises}, the negative log-likelihood for $\MRij$ is: 
\begin{equation}
\label{eq:nllR}
\begin{split}
- \log \calL( \MRij | \vxx) &= - \omegaR \; \trace{ \MR_j\tran \MR_i \MRij } + \mbox{const.} \\
&= \frac{\omegaR}{2} \normsq{ \MR_j - \MR_i \MRij}{\frob} + \mbox{const.}
\end{split}
\end{equation}
%
%
%
where $\normsq{\cdot }{\frob}$ is the Frobenius matrix norm (sum of the squares of the entries), 
and we used $\normsq{\MS - \MR}{\frob} \!=\! \trace{(\MS \!-\! \MR)(\MS \!-\! \MR)\tran}$. 
The norm~$\normsq{\MS - \MR}{\frob}$ is usually referred to as the \emph{chordal distance} between two 
rotations $\MS$ and $\MR$~\cite{Hartley13ijcv}.

Plugging~\eqref{eq:nllt} and \eqref{eq:nllR} back into~\eqref{eq:MLE} we obtain 
our \ML estimator:

\vspace{-0.4cm}
\bea
\fMLstar =  
\min_{\!\!\! \substack{ \{ \vt_i \in \Real{3} \} \\ \{ \MR_i \in \SOthree  \} }}  \hspace{0.2cm} 
 \sumalledges \hspace{0.2cm} \omegat \normsq{ \vt_j \!-\! \vt_i \!-\! \MR_i \vtij }{}  \hspace{1.0cm} \text{(\PGO)} \nonumber
\eea
\vspace{-0.6cm}
\bea
\hspace{3cm}  + \frac{\omegaR}{2} \normsq{ \MR_j \!-\! \MR_i \MRij}{\frob}  
\vspace{-0.1cm} \label{eq:PGO}
\eea

The main difference between~\eqref{eq:PGO} 
and formulations in related work is the use of the chordal distance 
(related work instead uses the \emph{geodesic distance} $\normsq{ \logmap{ \MRij\tran \MR_i\tran \MR_j } }{}$).   
References~\cite{Carlone15icra-verify,Hartley13ijcv} show that  
for small residual errors $\frac{1}{2} \normsq{ \MR_j - \MR_i \MRij}{\frob} \approx \normsq{ \logmap{ \MRij\tran \MR_i\tran \MR_j } }{}$, 
making the formulations equivalent from a practical standpoint.

The advantage of the formulation~\eqref{eq:PGO} is that it has a quadratic objective function. 
This facilitates the derivation of the Lagrangian dual problem, as shown in the 
next section.



\section{Lagrangian Duality in 3D PGO}
\label{sec:duality}

The main goal of this paper is to provide tools to check if a candidate SLAM solution $\candidate$
is globally optimal. If we knew the optimal cost $\fMLstar$ this
would be easy: calling $\fML(\cdot)$ the objective function of~\eqref{eq:PGO}, 
 if $\fML(\candidate)=\fMLstar$ then $\candidate$ is 
optimal. 
Unfortunately, $\fMLstar$ is unknown. Our contribution is to show that we can
compute close proxies of $\fMLstar$ using duality theory.
To make the derivation easier, we first rewrite the problem 
as a quadratic problem with equality constraints (Section~\ref{sec:QPQEC}), 
and then derive the dual (Section~\ref{sec:dual}).

\subsection{Quadratic Problem with Quadratic Equality Constrains}
\label{sec:QPQEC}

In this section, we rewrite~\eqref{eq:PGO} in order to 
(i) have vector variables (the rotations $\MR_i$ are matrices), and (ii) 
formulate the constraints $\MR_i\in\SOthree$ as quadratic equality constraints.

We define $\vr_i \in \Real{9}$ as the vectorized version of $\MR_i$: 
$\vr_i \doteq [\MR_i\at{1} \; \MR_i\at{2} \;\MR_i\at{3} ]\tran$, where $\MR_i\at{k}$ is the 
$k$th \emph{row} of $\MR_i$. We use the shorthand $\vr_i = \rows(\MR_i)$ to 
obtain the vector representation $\vr_i$ of a $3\times 3$ matrix $\MR_i$.
Using this parametrization, each summand in the objective in~\eqref{eq:PGO} becomes 
(using that $\|\MR\|_\frob = \|\MR\tran\|_\frob$ in the first expression):
\bea
\!\!\!\!& & \omegat \normsq{ \vt_j \!-\! \vt_i \!-\! \MR_i \vtij }{} + \frac{\omegaR}{2} \normsq{ \MR_j\tran \!-\! \MRij\tran \MR_i\tran}{\frob} \nonumber\\
\!\!\!\!&=& 
\omegat \normsq{ \vt_j \!-\! \vt_i \!-\! \vtijlift \vr_i  }{} + \frac{\omegaR}{2} \normsq{ \vr_j \!-\! \MRijlift \vr_i }{}
\!\!\!\!\!\! \label{eq:ObjVect}
\eea
where $\vtijlift \doteq \eye_3 \otimes \vtij\tran \in \Real{3 \times 9}$, 
$\MRijlift \doteq \eye_3 \otimes \MRij\tran \in \Real{9 \times 9}$, and 
 $\otimes$ is the Kronecker product.

We cannot choose arbitrary vectors  $\vr_i \in \Real{9}$, but have to limit ourself to 
choices of $\vr_i$ that produce meaningful rows of a rotation matrix $\MR_i \in \SOthree$. 
The rotation group $\SOthree$ is defined as $\SOthree \doteq \setdef{\MR \in \Real{3 \times 3}}{\MR\tran \MR = \eye_3, \det(\MR)=1}$, 
which, written in terms of the rows of $\MR_i$, becomes:
\bea
\label{eq:SO3constains}
\MR_i\tran \MR_i = \eye_3 \!\!&\!\!\Leftrightarrow \!\!&\!\! (\MR_i\at{u})\tran \MR_i\at{v} = 
\left\{
\ba{ll}
1 &  \text{if } u = v,    \\ 
0 &  \text{if } u \neq v, 
\ea
\right. 
\ba{l}
u,v=  \\ 
1,2,3
\ea
\nonumber \\
\det(\MR_i)=1
\!\!&\!\!\Leftrightarrow \!\!&\!\! \MR_i\at{1} \times  \MR_i\at{2} = \MR_i\at{3}
\eea
where $\times$ is the cross product. In other words, the rows of a rotation matrix 
have to be orthonormal, and have to satisfy the right-hand rule. 
To derive the dual problem we relax the second condition ($\det(\MR)=1$), which
 amounts to performing estimation in \Othree rather than \SOthree (i.e., resulting 
 matrices can have determinant $\det(\MR)=\pm 1$). Then in Proposition~\ref{prop:verificationTechniques} 
 we show how to reconcile our verification techniques to 
 work directly on the 
 original \PGO problem~\eqref{eq:PGO}.
 

Using~\eqref{eq:ObjVect} and~\eqref{eq:SO3constains} and relaxing the determinant constraints,
we rewrite the \PGO problem~\eqref{eq:PGO} as: 
%
\bea
\label{eq:PGOvect}
f^\star = \min_{ \{ \vr_i,\vt_i \} } \!\!\! &\!\!\!\!
\sumalledges \!\! \omegat \normsq{ \vt_j \!-\! \vt_i \!-\! \vtijlift \vr_i  }{} + \frac{\omegaR}{2} \normsq{ \vr_j \!-\! \MRijlift \vr_i }{} \nonumber \hspace{-0.3cm} \\
\subject & \hspace{-0.7cm}
\left. 
\ba{ll} 
\vr_i\tran \ME_{uv} \vr_i = 1, \quad u = v \qquad \\
\vr_i\tran \ME_{uv} \vr_i = 0, \quad u \neq v
\ea
\right\} 
\ba{ll} 
u,v=1,2,3  \\
i=1,\ldots,\nrNodes
\ea
\eea
where $\ME_{uv}$ is a $9\times 9$ selection matrix composed of $3\times 3$ blocks that are zero everywhere except the 
$3 \times 3$ block in position $(u,v)$, which is the identity matrix. The matrices $\ME_{uv}$ are built such that $\vr_i\tran \ME_{uv} \vr_i = 
(\MR_i\at{u})\tran \MR_i\at{v}$, hence the constraints in~\eqref{eq:PGOvect} correspond to the orthonormality
constraints in~\eqref{eq:SO3constains}. 

In order to write~\eqref{eq:PGOvect} in a more compact matrix notation, we define the 
vector $\breve\vxx = [\vt_1\tran, \ldots, \vt_\nrNodes\tran , \vr_1\tran, \ldots, \vr_\nrNodes\tran]\tran \in \Real{12\nrNodes}$. 
Using this notation,~\eqref{eq:PGOvect} becomes: 
\bea
\label{eq:PGOvect2}
f^\star = \min_{ \breve\vxx  }  &
\| \MAlift \breve\vxx \|^2 \hspace{5cm} \\
\subject & 
\left. 
\ba{ll} 
\breve\vxx\tran \MElift_{iuv} \breve\vxx = 1, \quad u = v \qquad \nonumber \\
\breve\vxx\tran \MElift_{iuv} \breve\vxx = 0, \quad u \neq v
\ea
\right\} 
\ba{ll} 
u,v=1,2,3 \nonumber \\
i=1,\ldots,\nrNodes
\ea
\eea
where the matrices $\MAlift$ and $\MElift_{iuv}$ 
are obtained by stacking the coefficient matrices in~\eqref{eq:PGOvect}, with  suitable  zero
blocks for padding; $\omegat$ and $\omegaR$ are included in the definition of $\MAlift$.
%

Finally, since absolute poses are not observable from relative measurements, we 
fix a pose to be our reference frame. 
Without loss of generality 
we fix the pose of the first node to the identity pose ($\vt_1 = \zero_3$ and $\MR_1 = \eye_3$, or, equivalently 
$\vr_1 = \rows(\eye_3)$). This process is usually called \emph{anchoring}. 
Fixing the first pose modifies~\eqref{eq:PGOvect2} as follows:
\bea
\label{eq:primalNonHom}
f^\star = \min_{ \vxx  }  &
\normsq{ \MAanc \vxx - \vb}{}  \hspace{4.5cm} \\
\subject & 
\left. 
\ba{ll} 
\vxx\tran \MEanc_{iuv} \vxx = 1, \quad u = v \qquad  \\
\vxx\tran \MEanc_{iuv} \vxx = 0, \quad u \neq v
\ea
\right\} 
\ba{ll} 
u,v=1,2,3 \\
i=\obsNodes 
\ea 
\nonumber \hspace{-0.3cm}
\eea
where $\vxx \in \Real{12(\nrObsNodes)}$ is obtained by removing the first pose from $\breve{\vxx}$, 
 $\MAanc$ is obtained by removing from $\MAlift$ the columns corresponding to the first pose, and 
 $\vb$ is the known right-hand-side arising from anchoring; $\MEanc_{iuv}$ 
 are the same as $\MElift_{iuv}$ but without the rows and columns corresponding to the first pose.

We conclude this section by transforming~\eqref{eq:primalNonHom} into an equivalent 
problem with \emph{homogeneous} objective
(i.e., without constant terms in the squared cost).
For this purpose, we note that solving~\eqref{eq:primalNonHom} is the same as 
solving:
\bea
\label{eq:primal}
f^\star = \min_{ \vxx, y  }  &
\normsq{ \MAanc \vxx - \vb y}{} \hspace{2cm} \text{(primal problem)} \nonumber \\
\subject & 
\left. 
\ba{ll} 
\vxx\tran \MEanc_{iuv} \vxx = 1, \quad u = v \qquad  \\
\vxx\tran \MEanc_{iuv} \vxx = 0, \quad u \neq v
\ea
\right\} 
\ba{ll} 
u,v=1,2,3 \\
i=\obsNodes
\ea  \nonumber \\
& y^2 = 1 \hspace{5cm}
\eea
meaning that the two problems have the same optimal objective, 
and the corresponding solutions can be mapped to each other.
Intuitively, if the solution of~\eqref{eq:primal} is $[\vxx_H^\star \; 1]$, 
then $\vxx^\star = \vxx_H^\star$ is also optimal for~\eqref{eq:primalNonHom}, while   
if the solution is $[\vxx_H^\star \; -1]$, then $\vxx^\star = -\vxx_H^\star$ will be 
optimal for~\eqref{eq:primalNonHom}.
The inclusion of the \emph{slack} variable $y$ is often referred to as \emph{homogenization}. 
We refer to~\eqref{eq:primal} as the \emph{primal} problem. 

%

\subsection{The dual problem}
\label{sec:dual}

In this section we apply Lagrangian duality to the primal problem~\eqref{eq:primal}, 
borrowing the corresponding theory from the optimization community~\cite{Boyd04book,Calafiore14book}. 
 We begin by recalling basic properties and notions about duality theory and then we 
tailor these concepts to our SLAM problem.

The key insight of duality is that for every constrained optimization problem of the form: 
\bea
\label{generic_equality_constrained_optimization_problem}
 f^\star =& \min_{\vxx} & f(\vxx) \\
 & \subject \; & c_i(\vxx) = 0 \quad \forall i \in \calC \nonumber
\eea
(where $\calC$ is a set indexing the constraints $c_i(\vxx)$), there is an associated \emph{unconstrained} optimization problem:
\begin{equation}
\label{Lagrangian_dual_problem_def}
d^\star = \max_{\vlambda} \bigg( \overbrace{\inf_{\vxx} f(\vxx) + \sum_{i \in \calC} \lambda_i c_i(\vxx)}^{d(\vlambda)} \bigg)
\end{equation}
called the \emph{dual problem}.  The scalar variables $\lambda_i$ appearing in \eqref{Lagrangian_dual_problem_def} are called \emph{Lagrange multipliers} 
or \emph{dual variables}, and the function $d(\vlambda)$ is called the \emph{dual function}; $\vlambda$ is a vector stacking all 
dual variables.
 With reference to the dual problem \eqref{Lagrangian_dual_problem_def}, problem \eqref{generic_equality_constrained_optimization_problem} is referred to as the \emph{primal} problem.
Intuitively, the minimization (``$\inf$'') in $d(\vlambda)$ can be understood as a relaxation of the original 
problem~\eqref{generic_equality_constrained_optimization_problem} in which the constraints 
are transformed into penalty terms in the objective, whose ``importance'' is controlled by $\vlambda$; 
hence, the maximization (w.r.t., $\vlambda$) tries to make this relaxation as 
tight as possible.

The dual problem \eqref{Lagrangian_dual_problem_def} has two important properties.  First, since the dual function $d(\vlambda)$ is the pointwise 
infimum of a family of affine functions of $\vlambda$, it is \emph{always} concave, and therefore the dual \emph{maximization} problem \eqref{Lagrangian_dual_problem_def} is a convex program \cite[Sec.\ 5.2]{Boyd04book}.  Its convexity guarantees that the dual problem can always be solved \emph{globally optimally} using \emph{local} search techniques.  Second, 
given any feasible $\vxx$ for \eqref{generic_equality_constrained_optimization_problem} (for which $c_i(\vxx) = 0$), the definition of $d(\vlambda)$ in \eqref{Lagrangian_dual_problem_def} shows that $d(\vlambda) \le f(\vxx)$ for any choice of $\vlambda$.  In particular, this must also hold at the optima $\vxx^\star$ and $\vlambda^\star$ in \eqref{generic_equality_constrained_optimization_problem} and \eqref{Lagrangian_dual_problem_def}, so that:
\begin{equation}
\label{eq:weakDuality}
 d^\star \le f^\star.
\end{equation}
The inequality \eqref{eq:weakDuality} is referred to as \emph{weak} (\emph{Lagrangian}) \emph{duality}, and it enables us to lower-bound the optimal value $f^\star$ of the (possibly very difficult, nonconvex) primal problem \eqref{generic_equality_constrained_optimization_problem} using the optimal value $d^\star$ of the (convex) dual problem \eqref{Lagrangian_dual_problem_def}.  For some problems, the inequality \eqref{eq:weakDuality} is tight (i.e.\ $d^\star = f^\star$), for which we say that \emph{strong} (\emph{Lagrangian}) \emph{duality} holds.  The quantity $f^\star - d^\star \ge 0$ is called the \emph{duality gap}. 

Using weak duality, it is easy to show that,
given a primal feasible point $\hat{\vxx}$ and a dual point $\hat{\vlambda}$, 
the following chain of inequality holds: 
\begin{equation}
\label{duality_certification_criteria}
 d(\hat{\vlambda}) \le d(\vlambda^\star) \doteq d^\star \le f^\star \doteq f(\vxx^\star) \le f(\hat{\vxx}) 
\end{equation}
where the first inequality stems from the fact that $\vlambda^\star$ attains the maximum over all 
$\vlambda$, and the last follows from the 
optimality of $f^\star$ (which is the global minimum among all feasible $\vxx$). 

Therefore, in this work we exploit a simple idea: given a candidate solution $\hat{\vxx}$, if we are 
able to find a $\hat{\vlambda}$, for which $d(\hat{\vlambda}) = f(\hat{\vxx})$, then the chain of inequalities~\eqref{duality_certification_criteria} 
becomes tight ($d(\hat{\vlambda}) = d^\star = f^\star =f(\hat{\vxx})$), which implies that $\hat{\vxx}$ is an optimal solution. 
Equation \eqref{duality_certification_criteria} thus provides a means of \emph{certifying} the global optimality of a candidate solution $\hat{\vxx}$ for \eqref{generic_equality_constrained_optimization_problem} 
and enables our derivation of \emph{algorithmic} approaches for certifying the correctness of SLAM solutions.

We are now ready to apply duality to our primal problem~\eqref{eq:primal}.
From \eqref{eq:primal} and \eqref{Lagrangian_dual_problem_def}, the dual function is:

%
\bea
\label{eq:dualFunction}
d(\vlambda) \!\!&=&\!\! \inf_{ \vxx, y  }\normsq{ \MAanc \vxx - \vb y}{} + 
\sumObsNodes \bigg[
\displaystyle \sum_{ \substack{ u=1,2,3} } \lamPara_{iuu} (1 - \vxx\tran \MEanc_{iuu} \vxx) \nonumber \\ 
&+& \!\!\!\! \displaystyle \sum_{ \substack{ u,v=1,2,3 \\ u \neq v} } \lamPerp_{iuv} (-\vxx\tran \MEanc_{iuv} \vxx )
\bigg]
+ \lamy (1 - y^2),
\eea
where $\vlambda$ is the vector of Lagrange multipliers $\lamPara_{iuv}$ and $\lamy$, associated with the orthonormality and  homogeneity constraints in \eqref{eq:primal}, respectively.
We observe that the quadratic terms in~\eqref{eq:dualFunction} can be written more compactly as:
\bea
\label{eq:quadraticTerms}
& \normsq{ \MAanc \vxx - \vb y}{} 
 - \vxx\tran \displaystyle \bigg[\sumObsNodes 
 \displaystyle \sum_{ \substack{ u,v=1} }^3 \!\!\! \lamPara_{iuv} \MEanc_{iuv}  \bigg] \vxx - \lamy \; y^2
 \nonumber  \\
=& \vect{\vxx \\ y}\tran 
\matTwo{\MH(\vlambda) & - \MAanc\tran \vb \\
- \vb\tran \MAanc &   \vb\tran \vb - \lamy} 
 \vect{\vxx \\ y},
\eea
where
\beq
\label{eq:MH}
\MH(\vlam) \doteq \MAanc\tran \MAanc - 
 \sumObsNodes 
\displaystyle \sum_{ \substack{ u,v=1} }^3 \!\!\! \lamPara_{iuv} \MEanc_{iuv}.
\eeq
Calling $\MM(\vlambda)$ the matrix in~\eqref{eq:quadraticTerms}, the 
dual function~\eqref{eq:dualFunction} can thus be written as:
\bea
\label{eq:dualFunctionCompact}
\hspace{-0.5cm}d(\vlambda) \!\!\!\!&=&\!\!\!\! \inf_{ \vxx, y  } \vect{\vxx \\ y}\tran \MM(\vlambda) \vect{\vxx \\ y} + 
\!\!\sum_{ \substack{i=\obsNodes  \\ u=1,2,3} } \!\!\!\!\! \lamPara_{iuu} + \lamy.
\eea
Now in the dual problem \eqref{Lagrangian_dual_problem_def} we try to maximize $d(\vlam)$; however, from~\eqref{eq:dualFunctionCompact} 
we see that $d(\vlam) = -\infty$ if $\MM(\vlam)$ has a negative eigenvalue (by letting $[\vxx, y]$ lie in the corresponding eigenspace).  Consequently,
 we can safely restrict our search to the vectors $\vlam$ that preserve positive 
 semi-definiteness of $\MM(\vlam)$~\cite[Sec.\ 5.1.5]{Boyd04book}.
 Moreover: 
 \beq
 \label{eq:nullSpacexy}
 \MM(\vlam)\succeq 0 \quad \Rightarrow \quad
 \inf_{\vxx,y}  \vect{ \vxx \\ y}\tran 
\MM(\vlam)
\vect{ \vxx \\ y} = 0, 
\eeq
as the minimization over $[\vxx \;y]$ in the homogenized problem ~\eqref{eq:nullSpacexy} 
is unconstrained.
The dual problem~\eqref{Lagrangian_dual_problem_def} thus becomes:
\bea
\label{eq:dualProblem}
d^\star =& \max_{\vlambda} & \textstyle \sum_{ \substack{i=1,\ldots,n \\ u=1,2,3} } \lamPara_{iuu} + \lamy \quad \quad \text{(dual problem)} \nonumber \\
 & \subject & \MM(\vlam) \succeq 0.
\eea
%
%
%
The dual SLAM problem turns out to be a \emph{semidefinite program} (SDP), for
which specialized solvers exist \cite{Yamashita2003Implementation}.   In the following section we discuss the 
  relations between the primal and the dual problem, and elucidate on its practical use. 
%

 


\section{Relation between the Primal and \\ the Dual Problem and Practical Use}
\label{sec:practicalUse}

In this section we present two powerful applications of the dual 
problem~\eqref{eq:dualProblem}. 
Section~\ref{sec:verification} deals with the case in which one 
is given a candidate \PGO solution, and wants to evaluate its quality, possibly 
certifying its optimality.
Section~\ref{sec:primalViaDual} shows that in particular cases 
(when the duality gap is zero) one can obtain an optimal solution 
of the primal problem from the solution $\vlambda^\star$ of the dual.

In both sections, we use the following property.

\begin{lemma}[Primal optimal solution and zero duality gap]
\label{lem:xstarInNullSpace}
If the duality gap is zero ($d^\star = f^\star$), then any 
primal optimal solution $[\vxx^\star \; 1]$ of~\eqref{eq:primal} is in the null space of the 
matrix $\MM(\vlambda^\star)$, where $\vlambda^\star$ is the solution of the dual problem~\eqref{eq:dualProblem}.
\end{lemma}

\begin{IEEEproof} 
When the duality gap is zero, any minimizer of the primal problem is also a minimizer for the infimum in the dual function \eqref{Lagrangian_dual_problem_def} \cite[Sec.\ 5.5.5]{Boyd04book}.
 Consider a primal optimal solution $[\vxx^\star \; 1]$. 
 We already observed than any such minimizer annihilates the quadratic term in~\eqref{eq:dualFunctionCompact}, and therefore
 it holds that $[\vxx^\star \; 1]\tran \MM(\vlamStar) [\vxx^\star \; 1] = 0$, 
 which implies that $[\vxx^\star \; 1]$ is in the null space of $\MM(\vlambda^\star)$, proving the claim.
 \end{IEEEproof} 

We give an alternative proof, which does not require 
prior knowledge on duality, in Appendix A.

\subsection{Verification}
\label{sec:verification}

In this section we consider the case in which we are given a 
candidate solution $\candidate$ for the primal problem, and we 
want to evaluate the quality of this solution. 
For brevity, we denote with $f(\candidate)$ the objective evaluated at $\candidate$, for
both~\eqref{eq:primalNonHom} and its homogeneous form~\eqref{eq:primal} (for the latter we imply $y=1$).

We begin with the following proposition, whose proof easily follows from~\eqref{duality_certification_criteria}  
and the discussion in Section \ref{sec:dual}.

\begin{proposition}[Verification of Primal Optimal Objective]
\label{prop:verifyOptimalObjective}
Given a candidate solution $\candidate$ for the primal problem~\eqref{eq:primal}, if $f(\candidate) = d^\star$, then the 
duality gap is zero and $\candidate$ is an optimal solution of~\eqref{eq:primal}. 
Moreover, even if the duality gap is nonzero, $f(\candidate) - d^\star \geq f(\candidate) - f^\star$, 
meaning that $f(\candidate) - d^\star$ is an upper-bound for the sub-optimality gap of $\candidate$.
\end{proposition}
%

\prettyref{prop:verifyOptimalObjective} ensures that the candidate $\candidate$ is optimal when $f(\candidate) = d^\star$. 
Moreover, even in the case in which we get $f(\candidate) > d^\star$, the quantity $f(\candidate) - d^\star$ can be 
used as an indicator of how far $\candidate$ is from the global optimum. 

While \prettyref{prop:verifyOptimalObjective} already provides means of verifying a candidate
solution, it requires solving the dual problem, to compute $d^\star$. The following proposition
 provides a technique to verify the optimality of $\candidate$ without solving the SDP. 

 \begin{proposition}[Verification of Primal Optimal Solution]
\label{prop:verifyOptimalSolution}
Given a candidate solution $\candidate$ for the primal problem~\eqref{eq:primal}, 
if the solution $\vlamHat$ of the linear system
\beq
\label{eq:verification2}
\MM(\hat{\vlambda}) \vect{\candidate \\ 1} = \zero  \qquad \text{(to be solved w.r.t. $\vlamHat$)}
\eeq
is such that $\MM(\hat{\vlambda})\succeq 0$ and $d(\vlamHat) = f(\candidate)$, then the
duality gap is zero and $\candidate$ is a primal optimal solution. 
\end{proposition}

\begin{IEEEproof} 
From~\prettyref{lem:xstarInNullSpace}, we know that when the duality gap is 
zero, it must hold $\MM(\vlamStar) [\vxx^\star \; 1] = \zero$. 
Therefore, in~\prettyref{prop:verifyOptimalSolution} we solve 
the linear system~\eqref{eq:verification2}, trying to obtain $\vlamStar$.
When $\MM(\hat{\vlambda})\succeq 0$, the solution $\vlamHat$ of~\eqref{eq:verification2} 
is such that $d(\vlamHat) \leq d^\star$ (recall that $d^\star$ is the maximum
 over $\vlambda$). Therefore, it holds that (i) $d(\vlamHat) \leq d^\star \leq f^\star \leq f(\candidate)$ 
 (by weak duality and optimality of $f^\star$). 
However, if $d(\vlamHat) = f(\candidate)$, the chain of inequalities 
(i) becomes tight, $d(\vlamHat) = d^\star = f^\star = f(\candidate)$, implying 
that $\candidate$ attains the optimal objective $f^\star$. 
 \end{IEEEproof} 

In practice, iterative SLAM solvers optimize~\eqref{eq:PGO}, rather than 
the primal problem~\eqref{eq:primal}. A natural question is then how to use 
the results in Propositions~\ref{prop:verifyOptimalObjective}-\ref{prop:verifyOptimalSolution} 
(which relate $f^\star$ and $d^\star$) to verify the solution of~\eqref{eq:PGO}, whose optimal value is 
$\fMLstar \geq f^\star$. 
This extension is given by the following proposition, which essentially states that 
Propositions~\ref{prop:verifyOptimalObjective}-\ref{prop:verifyOptimalSolution} 
can be applied directly to check the solution of~\eqref{eq:PGO}. 

\begin{proposition}[Verification techniques for PGO]
\label{prop:verificationTechniques}
The following statements hold true:
\begin{enumerate}[label=({\tt V}\arabic*)]
\item Given a candidate solution $\candidate$ for the \PGO problem~\eqref{eq:PGO}, 
if $\fML(\candidate) = d^\star$, then $\candidate$ is an optimal solution of~\eqref{eq:PGO}. 
Moreover, $\fML(\candidate) - d^\star \geq \fML(\candidate) - \fMLstar$, 
i.e., $\fML(\candidate) - d^\star$ is an upper-bound for the sub-optimality gap of $\candidate$.

\item Given a candidate solution $\candidate$ for the \PGO problem~\eqref{eq:PGO},
if the solution $\vlamHat$ of the linear system~\eqref{eq:verification2} is such that 
$\MM(\hat{\vlambda})\succeq 0$ and $d(\vlamHat) = \fML(\candidate)$, then the
duality gap is zero and $\candidate$ is an optimal solution of~\eqref{eq:PGO}.

\end{enumerate}
\end{proposition}

\begin{IEEEproof} 
The first claim can be proven by observing that the following chain of inequalities holds
 (i) $d^\star \leq f^\star \leq \fMLstar \leq \fML(\candidate)$, hence $\fML(\candidate) = d^\star$ 
 implies that $d^\star = f^\star = \fMLstar = \fML(\candidate)$, which implies that $\candidate$ 
 is optimal. The inequality $\fML(\candidate) - d^\star \geq \fML(\candidate) - \fMLstar$ 
 easily follows from (i). The second claim can be proven in the same way, noting that 
 (ii) $d(\vlamHat) \leq d^\star \leq f^\star \leq \fMLstar \leq \fML(\candidate)$: if we are able to compute 
 a $\vlamHat$ that is dual feasible ($\MM(\hat{\vlambda})\succeq 0$) and such that 
 $d(\vlamHat) = \fML(\candidate)$, then it must hold  $d(\vlamHat) = d^\star = f^\star = \fMLstar = \fML(\candidate)$, 
 which implies that $\fML(\candidate)$ is optimal.
 \end{IEEEproof} 


\newcommand{\mySpace}{\hspace{-0.2cm}}
\newcommand{\myFont}[1]{\hspace{0.4cm}{\small #1}}

\begin{figure*}[t]
\begin{minipage}{\textwidth}
\begin{tabular}{ccccc}%
\begin{sideways}{$\!\!\!\!\!\!\!\!\!\!\!${Verification 1}}\end{sideways} 
& \mySpace
\begin{minipage}{4cm}%
\centering%
\includegraphics[scale=0.24]{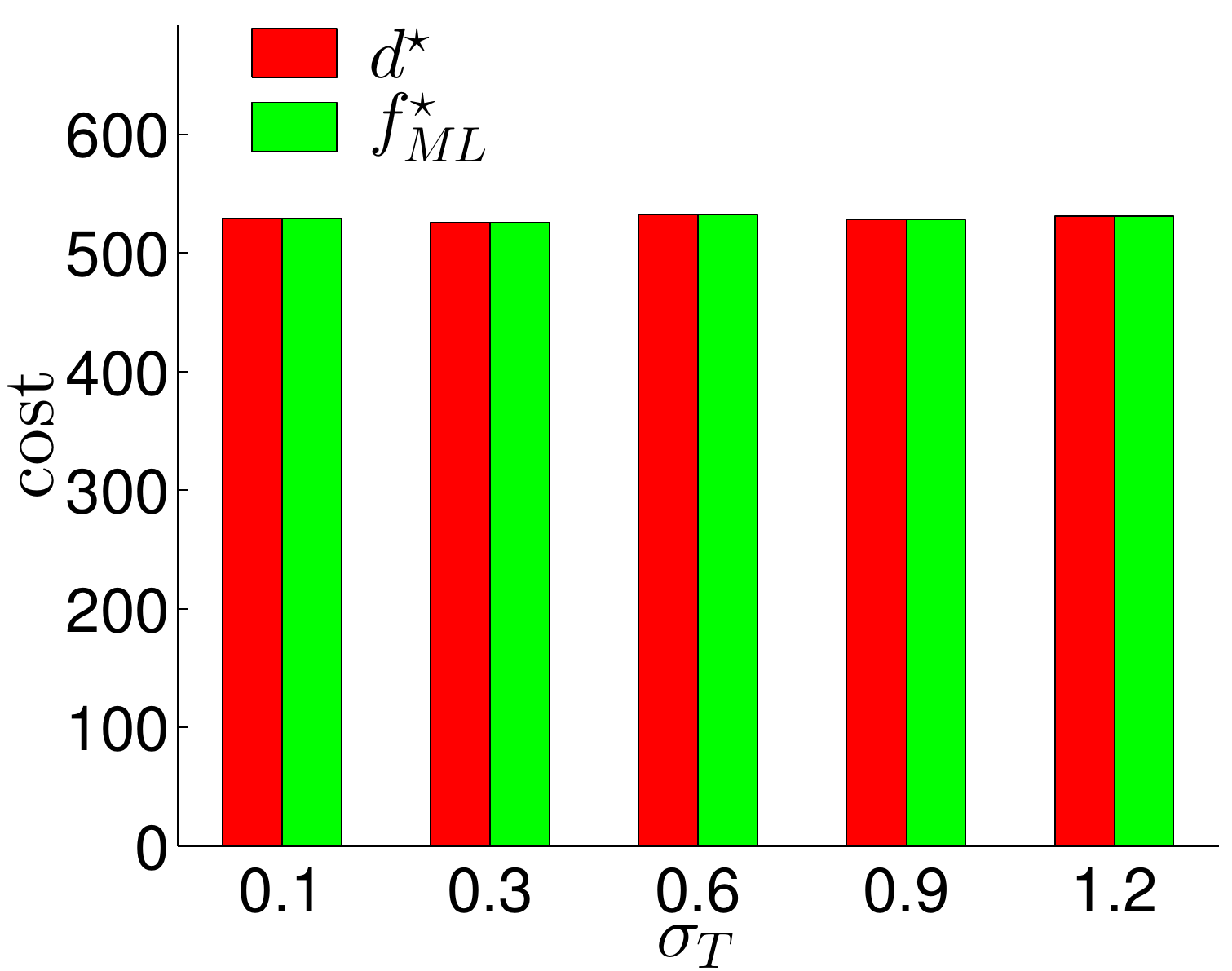} \\
\myFont{(a1)}
\end{minipage}
& \mySpace 
\begin{minipage}{4cm}%
\centering%
\includegraphics[scale=0.24]{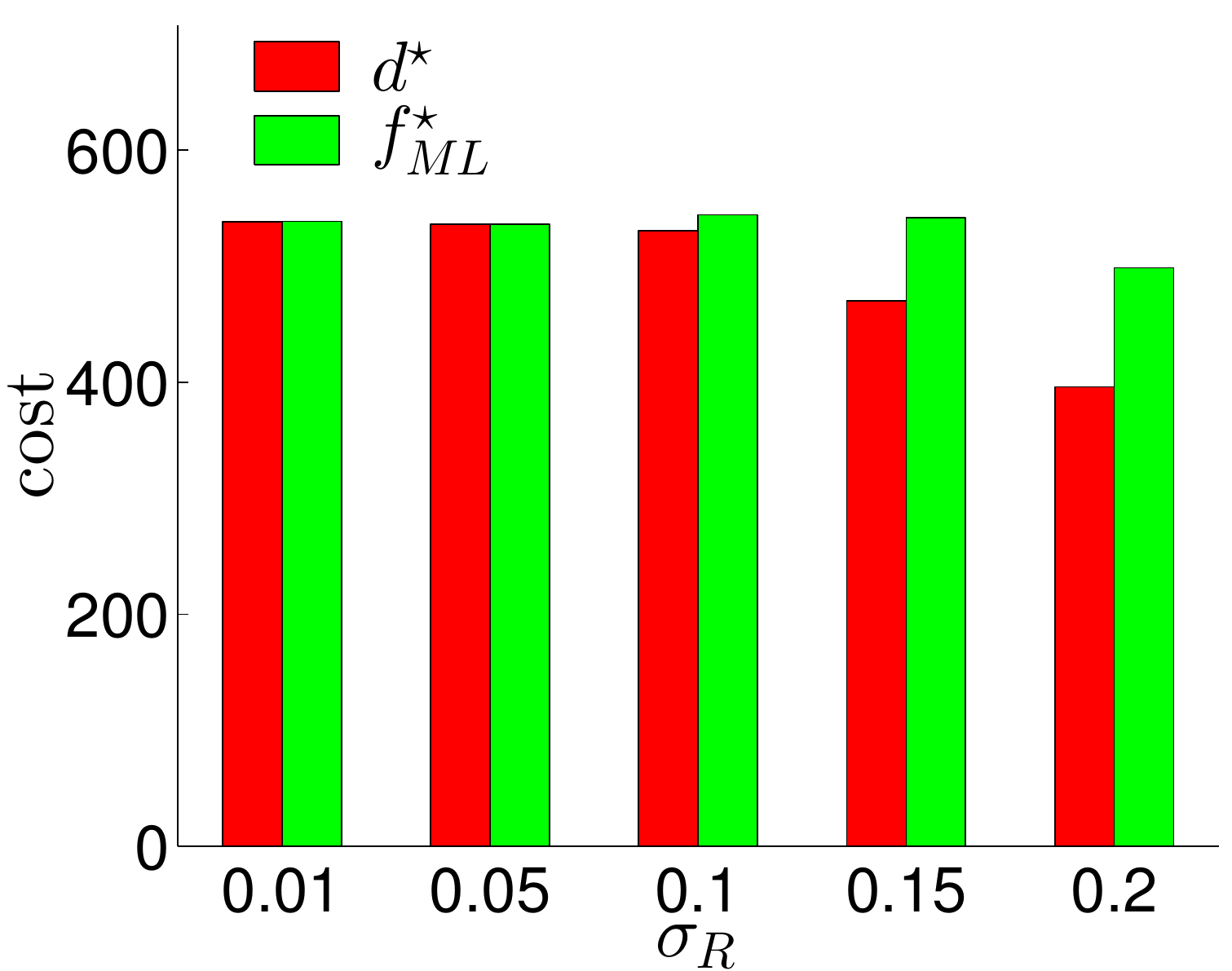} \\
\myFont{(a2)}
\end{minipage}
& \mySpace
\begin{minipage}{4cm}%
\centering %
\includegraphics[scale=0.24]{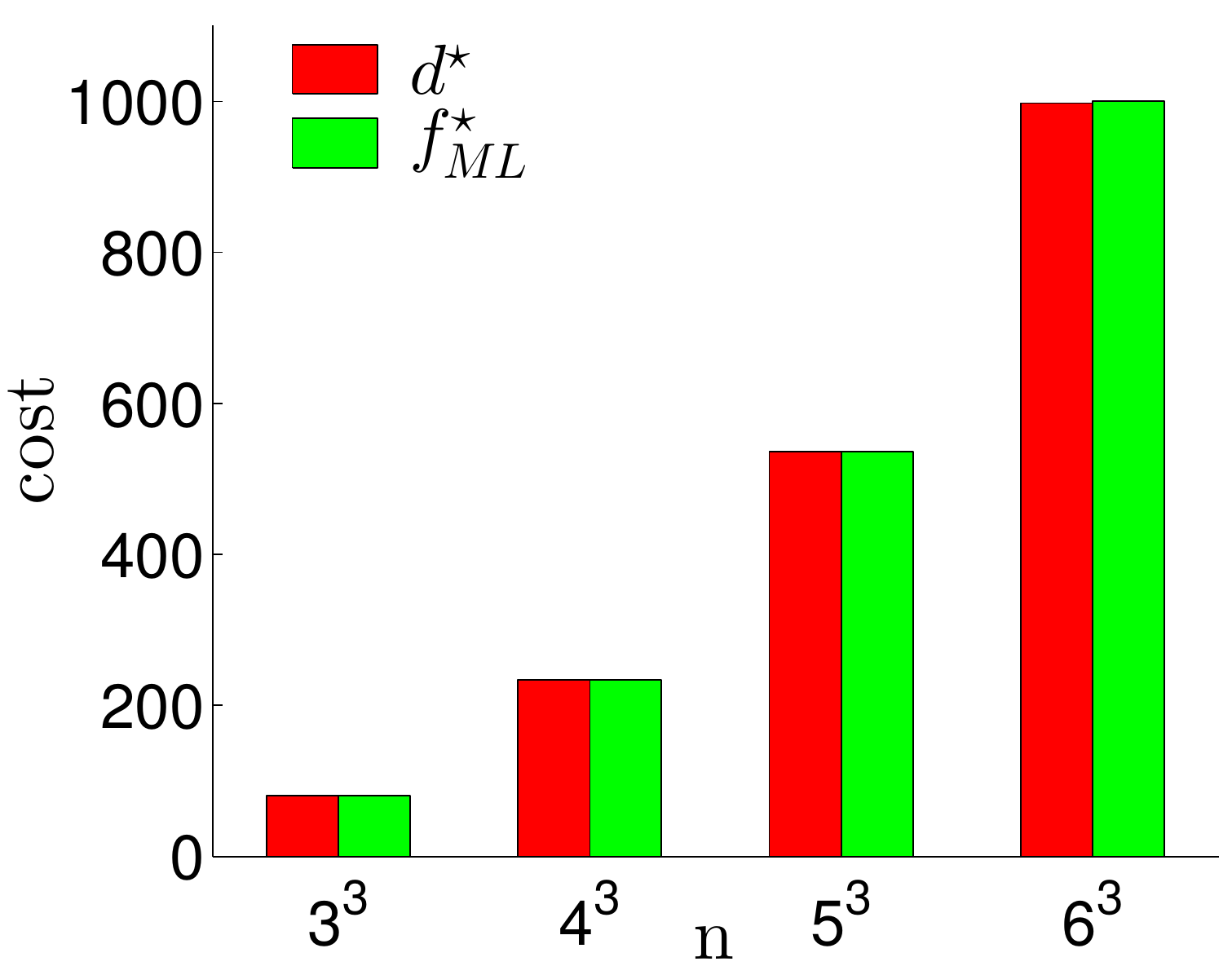} \\
\myFont{(a3)}
\end{minipage}
& \mySpace
\begin{minipage}{4cm}%
\centering%
\includegraphics[scale=0.24]{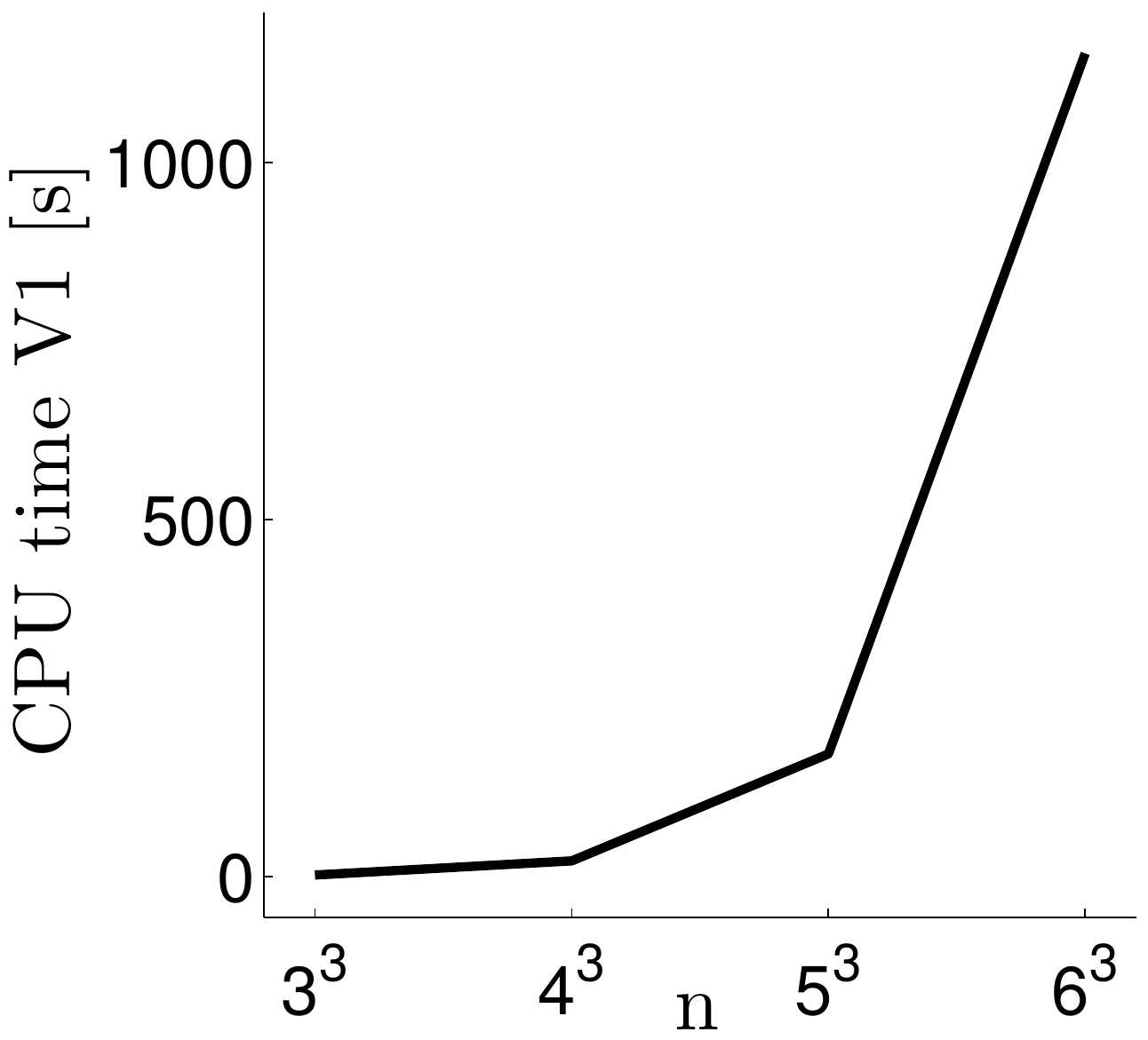} \\
\myFont{(a4)}
\end{minipage}
\\
\begin{sideways}{$\!\!\!\!\!\!\!\!\!\!\!${Primal via dual}}\end{sideways} 
&  \mySpace
\begin{minipage}{4cm}%
\centering%
\includegraphics[scale=0.23]{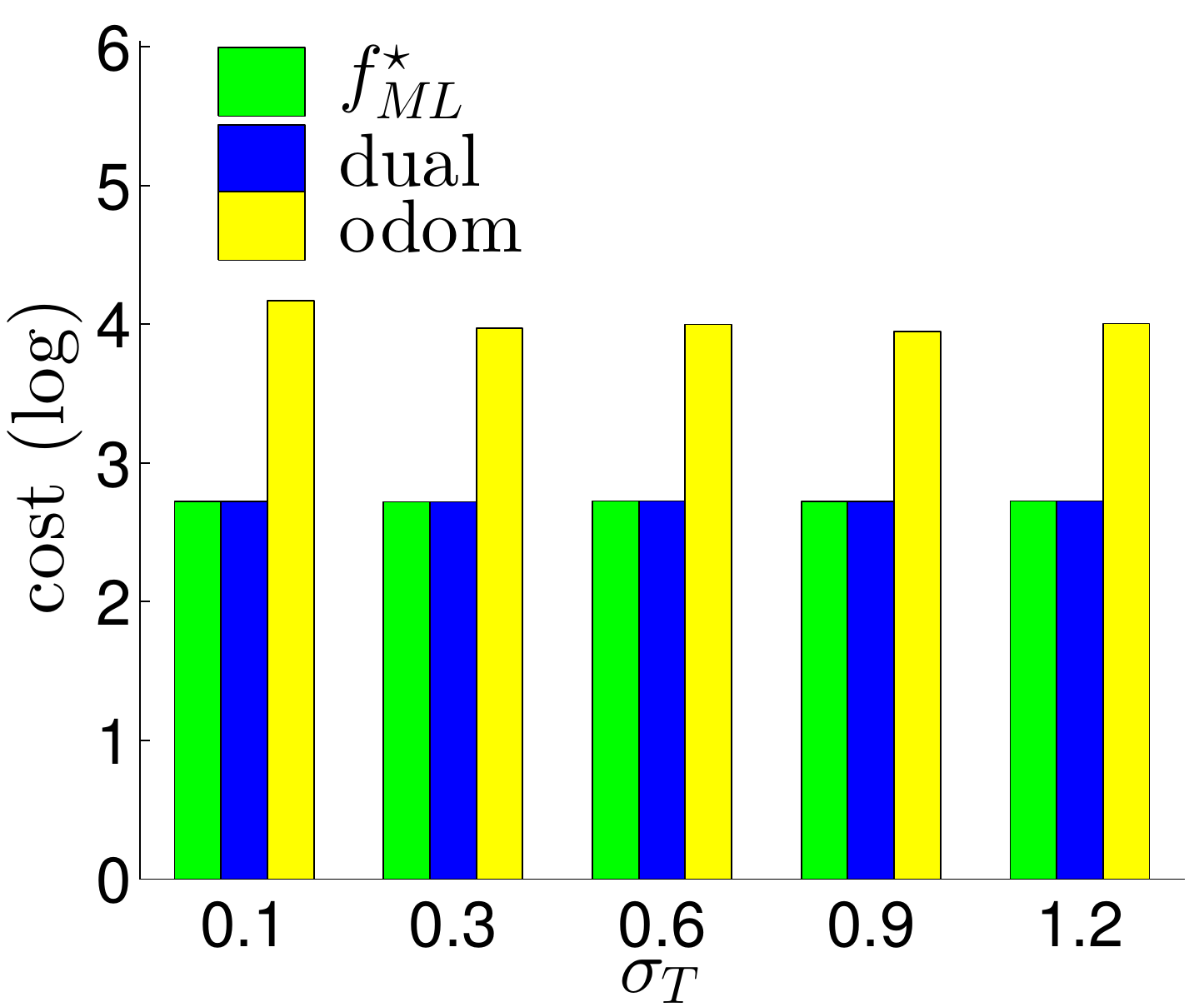} \\
\myFont{(b1)}
\end{minipage}
& \mySpace
\begin{minipage}{4cm}%
\centering%
\includegraphics[scale=0.23]{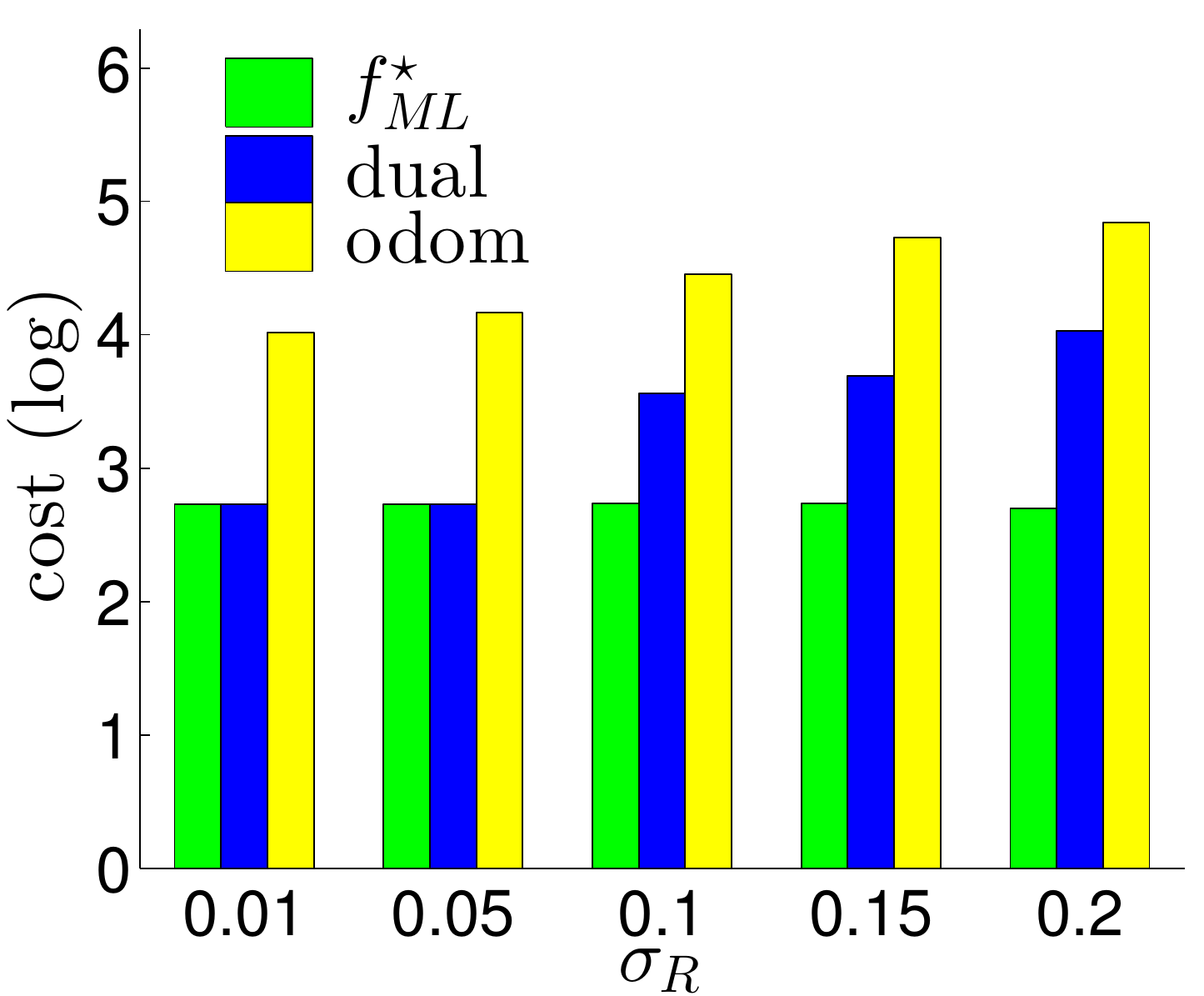} \\
\myFont{(b2)}
\end{minipage}
& \mySpace
\begin{minipage}{4cm}%
\centering %
\includegraphics[scale=0.23]{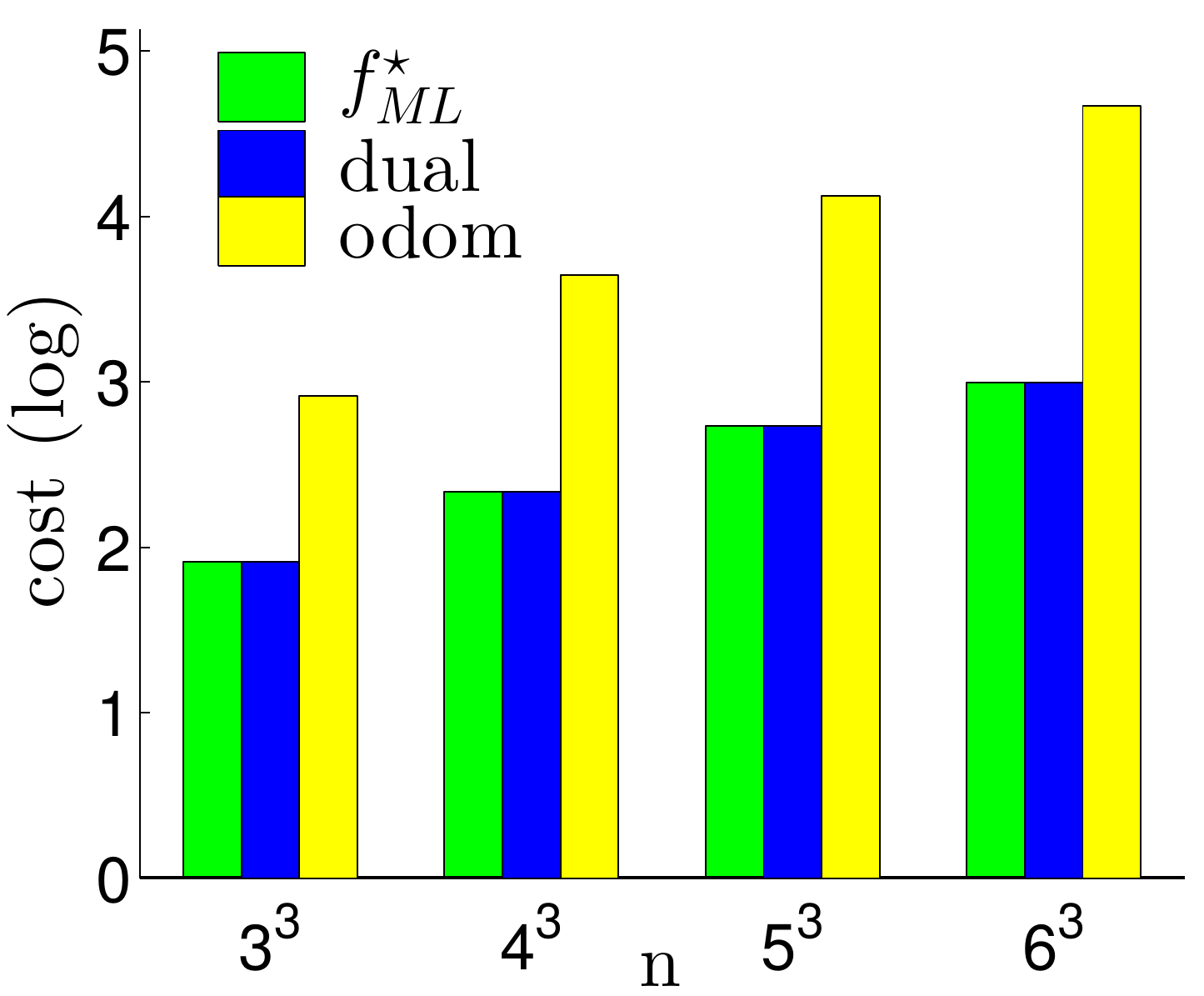} \\
\myFont{(b3)}
\end{minipage}
& \mySpace
\begin{minipage}{4cm}%
\centering%
\includegraphics[scale=0.29, trim = 0mm 20mm 0mm 0mm,clip]{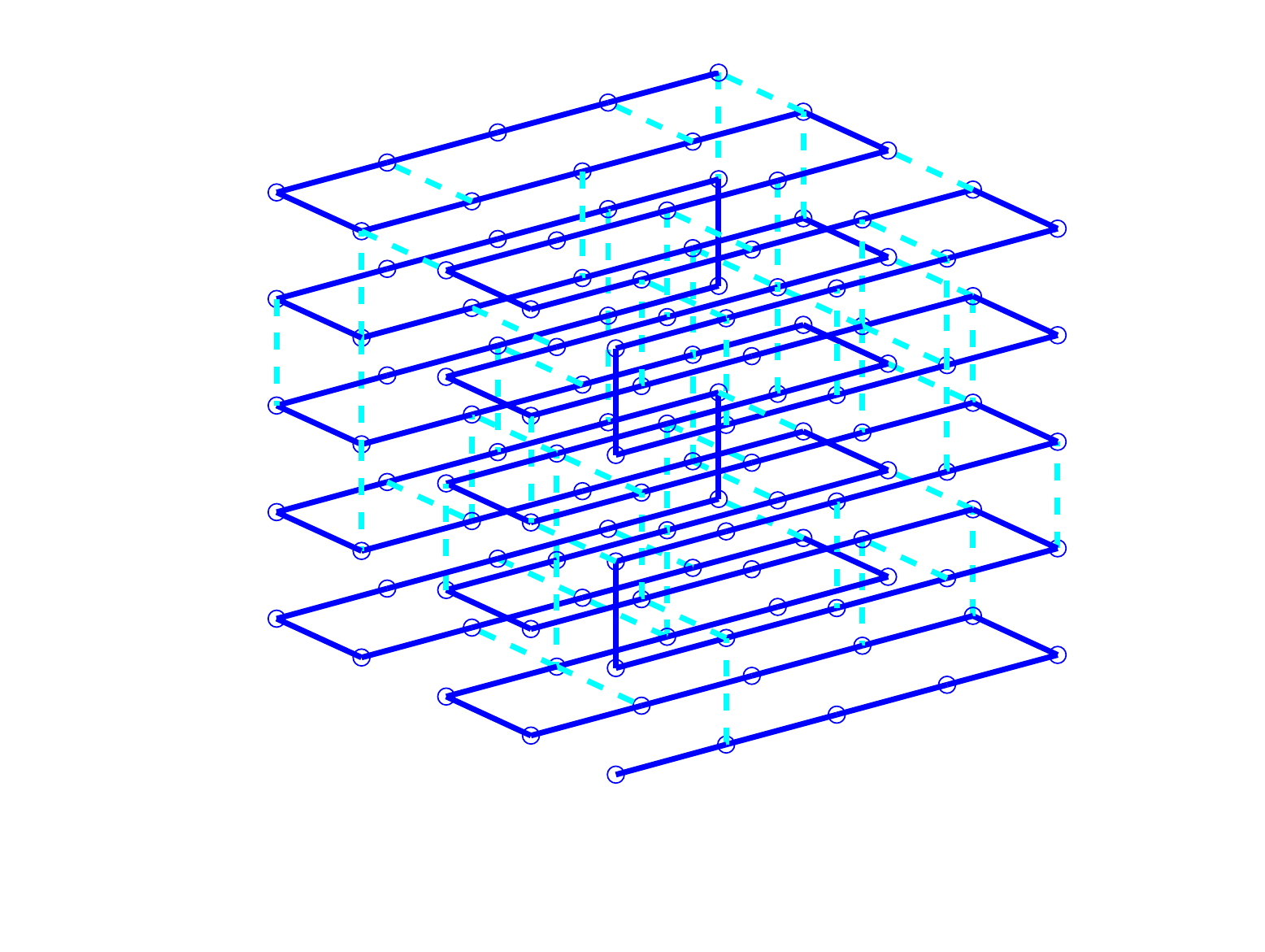} \\ 
\myFont{(b4)}
\end{minipage}
\\
\begin{sideways}{$\!\!\!\!\!\!\!\!\!\!\!${Verification 2}}\end{sideways} 
& \mySpace
\begin{minipage}{4cm}%
\centering%
\includegraphics[scale=0.25]{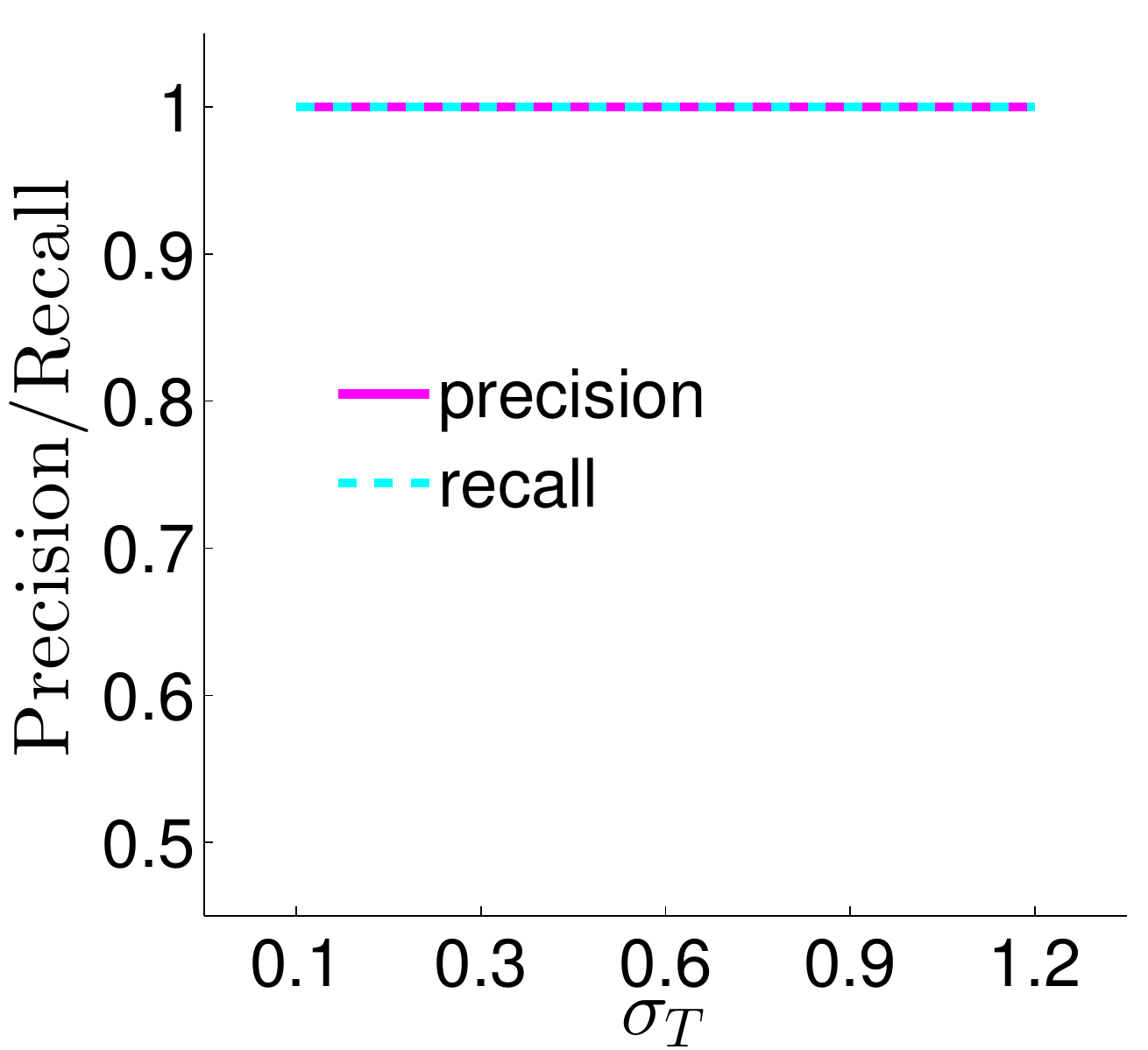} \\
\myFont{(c1)}
\end{minipage}
& \mySpace
\begin{minipage}{4cm}%
\centering%
\includegraphics[scale=0.25]{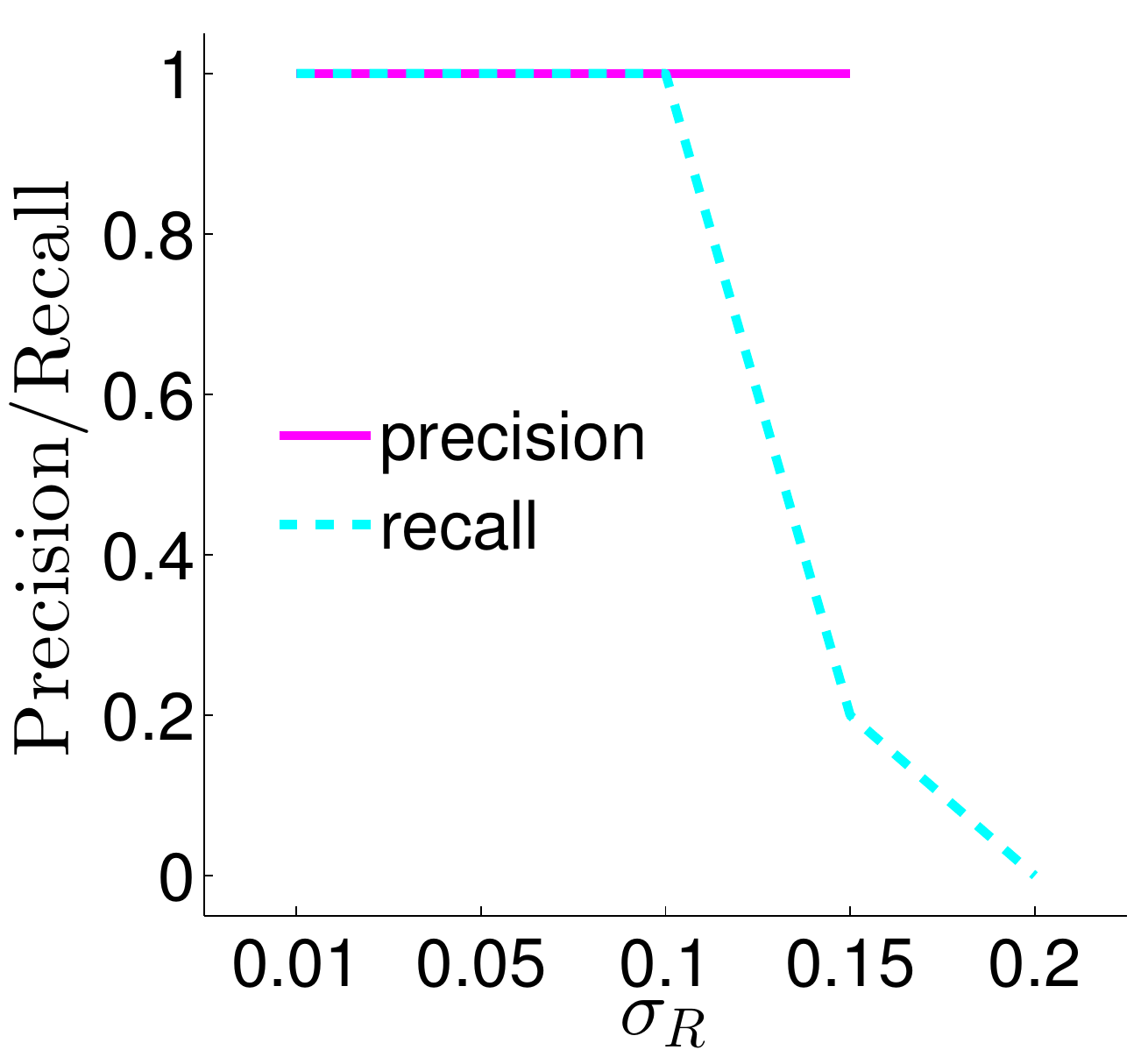} \\
\myFont{(c2)}
\end{minipage}
& \mySpace
\begin{minipage}{4cm}%
\centering %
\includegraphics[scale=0.25]{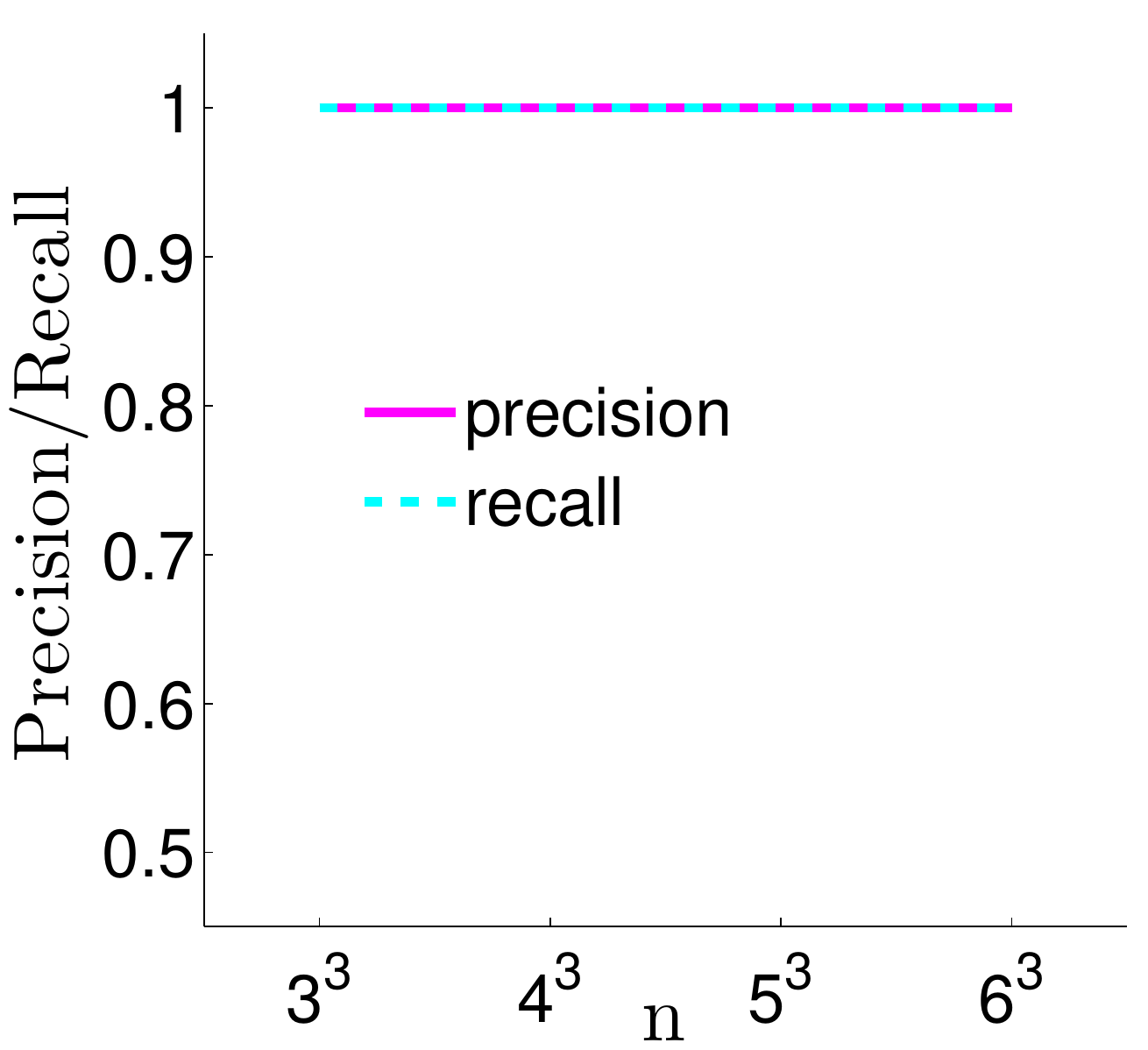} \\ 
\myFont{(c3)}
\end{minipage}
& \mySpace
\begin{minipage}{4cm}%
\centering%
\includegraphics[scale=0.25]{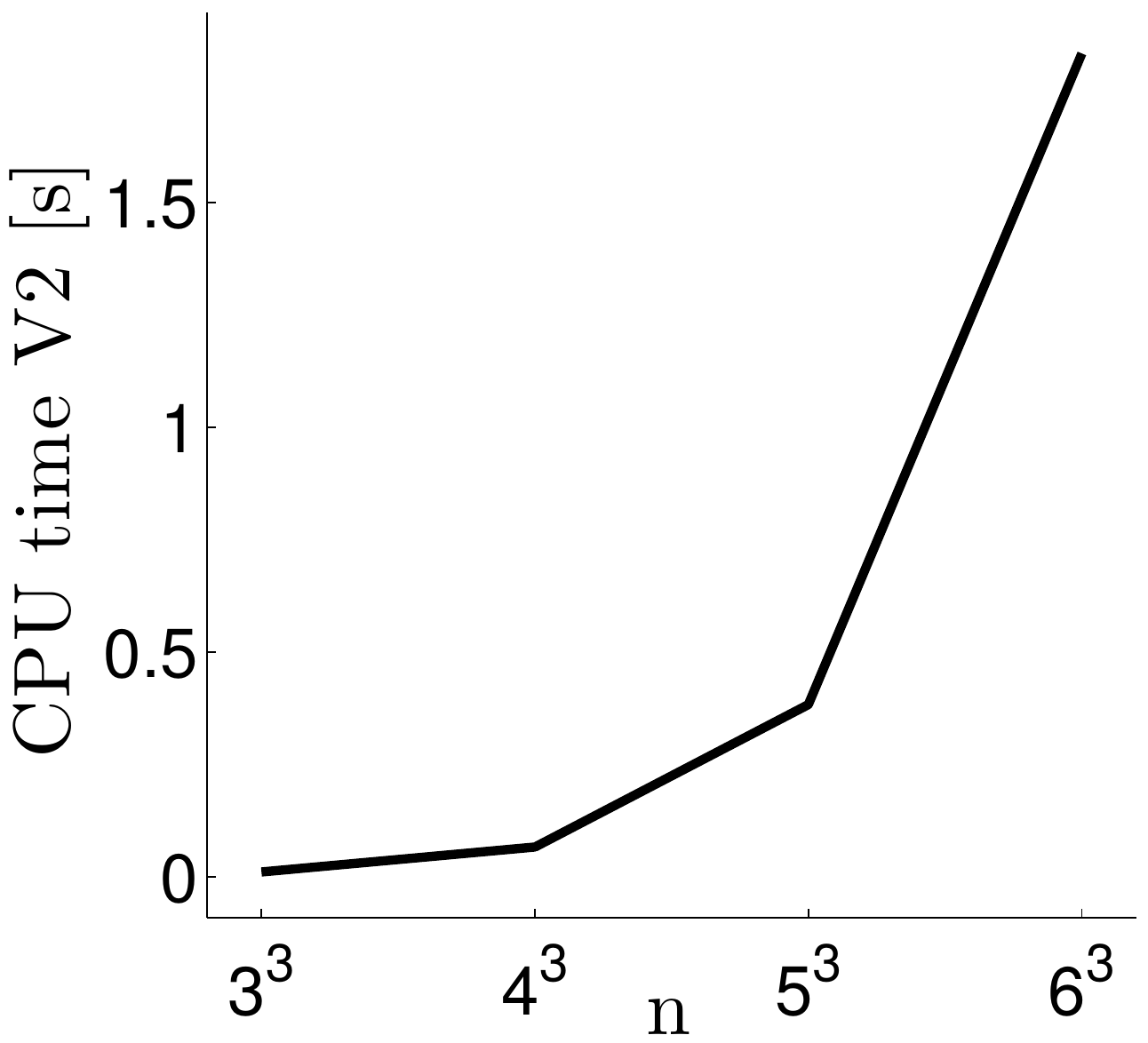} \\ 
\myFont{(c4)}
\end{minipage}
\end{tabular}%
\end{minipage}%
\caption{\label{fig:joint}
Statistics regarding the verification technique \verificationOne (first row), the computation of the primal solution via the dual
(second row), and the verification technique \verificationTwo (third row). Results are shown for different levels 
of translation noise $\sigmaT$ (first column), rotation noise $\sigmaR$ (second column), and size of the problem (third column).
The rightmost column shows the CPU time required by  \verificationOne and \verificationTwo and the datasets used for 
the Monte Carlo simulations.
}
\vspace{-0.5cm}
\end{figure*}

\subsection{Primal Optimal Solutions}
\label{sec:primalViaDual}

In this section, rather than verifying the quality of a given candidate solution, 
we show how to use duality to compute a primal optimal solution directly.
We focus on the particular case in which the duality gap is zero. 
This case is of interest as we observe that the duality gap 
is often zero in practice.

From~\prettyref{lem:xstarInNullSpace} we know that an optimal solution 
must be in the null space of $\MM(\vlamStar)$. This motivates the following proposition, 
which provides a way to compute a primal optimal solution directly
from the solution $\vlamStar$ of~\eqref{eq:dualProblem}.

\begin{proposition}
\label{prop:primalOptimalViaDual}
If the duality gap is zero and $\vlamStar$ is an optimal solution of~\eqref{eq:dualProblem}, 
then an optimal solution $\vxxStar$ of~\eqref{eq:primal} can be computed by solving the following linear system:
\beq
\label{eq:primalSolutionComputation}
\vect{\MH(\vlamStar) \\ \vb\tran \MAanc} \vxxStar = 
\vect{\MAanc\tran \vb \\ \lamyStar - \vb\tran \vb} \quad \text{(to be solved w.r.t. $\vxxStar$)}
\eeq
\end{proposition}
\vspace{0.2cm}

\begin{IEEEproof} 
From~\prettyref{lem:xstarInNullSpace}, we know that when the duality gap is 
zero, it holds $\MM(\vlamStar) [\vxxStar \; 1] = \zero$. Then, recalling the 
structure of $\MM(\vlamStar)$ from~\eqref{eq:quadraticTerms}, it's easy to 
see that~\eqref{eq:primalSolutionComputation} only rewrites the condition 
$\MM(\vlamStar) [\vxxStar \; 1] = \zero$, moving the constant terms to the 
right-hand side.
 \end{IEEEproof}

Note that \prettyref{lem:xstarInNullSpace} ensures that the linear system~\eqref{eq:primalSolutionComputation}
admits a solution. Proposition~\ref{prop:primalOptimalViaDual} allows finding an optimal 
solution to the primal problem~\eqref{eq:primal}. If this satisfies the 
determinant constraints in~\eqref{eq:PGO}, it follows that this solution 
is also optimal for the original \PGO problem~\eqref{eq:PGO}.
While currently we cannot prove that the determinant constraints 
are always satisfied, this was always the case in our experiments.

%
%
%




\newcommand{\GN}{{\tt GN}\xspace}
\newcommand{\accepted}{\green{\checkmark}\;}
\newcommand{\rejected}{\red{\mbox{X}}\;}
\newcommand{\red}[1]{{\color{red}#1}}
\newcommand{\green}[1]{{\color{green}#1}}

\section{Experiments}
\label{sec:experiments}

In this section we show that the first verification technique in \prettyref{prop:verificationTechniques}  
(referred to as \verificationOne)
enables accurate quantification of the sub-optimality gap of a candidate SLAM
solution, but that current SDP solvers do not scale well with increasing problem size.
The second verification technique in \prettyref{prop:verificationTechniques}
(referred to as \verificationTwo)
provides a convenient alternative for large-scale problems: it is reliable and less computationally demanding.
Finally, we provide empirical evidence that when the duality gap is zero we can compute an optimal solution 
from the solution of the dual problem, as suggested by \prettyref{prop:primalOptimalViaDual}.  
The SDP~\eqref{eq:dualProblem} is solved using \SDPA~\cite{Yamashita2003Implementation}. 
%

{\bf Effectiveness of \verificationOne.}  
\prettyref{prop:verificationTechniques}(\verificationOne) ensures that for any candidate solution $\candidate$,
$\fML(\candidate)-d^\star$ represents an upper-bound on the sub-optimality gap $\fML(\candidate) - \fMLstar$ of $\candidate$. However, it is possible that this bound
 is very loose, in which case it would be of little practical utility. 
In this section, we show that $d^\star$ is close to $\fMLstar$   
in practice, hence $\fML(\candidate)-d^\star$ is a very good measure of the sub-optimality gap of the candidate $\candidate$. 

 In our experiments we compute the ``optimal'' solution $\fMLstar$ of~\eqref{eq:PGO} 
 by refining the \emph{chordal} initialization of~\cite{Carlone15icra-init3D,Martinec07cvpr} 
 with 10 Gauss-Newton (\GN) iterations.
 While one cannot guarantee \emph{a priori}
 that this approach always produces the optimal estimate, using 
 the results of this paper we will be able to check optimality a posteriori.

We evaluated how close $d^\star$ is to $\fMLstar$ on the \grid dataset of Fig.~\ref{fig:joint}(b4). 
In this dataset, the odometric trajectory is simulated as the robot travels on a 3D grid world, 
and random loop closures are added between nearby nodes, with probability $0.3$.
Relative pose measurements are obtained by contaminating the true relative poses 
with zero-mean Gaussian noise, with standard deviation $\sigmaT$ and $\sigmaR$ for the translational and rotational noise, respectively.
Statistics are computed over 10 runs: 
for each run we create a \grid with random connectivity and random measurement noise. We consider an example with $n=5^3$ poses and varying noise levels $\sigmaT$ and $\sigmaR$. 

Fig.~\ref{fig:joint}(a1) shows $d^\star$ and $\fMLstar$ for different translational noise levels, fixing $\sigmaR = 0.05\mbox{rad}$.
The figure shows that $d^\star = \fMLstar$ (zero duality gap) independently on the translational noise level, hence 
$d^\star$ is a very good proxy of $\fMLstar$.

Fig.~\ref{fig:joint}(a2) shows $d^\star$ and $\fMLstar$ for different rotational noise, fixing $\sigmaT = 0.1\mbox{m}$.
In this case the duality gap $\fMLstar - d^\star$ is more sensitive to the noise level, and for large rotational noise 
$d^\star$ becomes smaller than $\fMLstar$. However, the gap $\fMLstar - d^\star$ remains small, 
and is within $20\%$ of $\fMLstar$ in all cases.

Fig.~\ref{fig:joint}(a3) shows $d^\star$ and $\fMLstar$ for different sizes of the \grid dataset, fixing $\sigmaR = 0.05\mbox{rad}$
and $\sigmaT = 0.1\mbox{m}$. Again in this case $d^\star = \fMLstar$ (zero duality gap) independently of the size of the dataset.
However, we observe that the SDP~\eqref{eq:dualProblem} becomes intractable for larger problem sizes, as shown in Fig.~\ref{fig:joint}(a4).
\journalVersion{While the interior-point algorithm used to solve the SDP 
has polynomial complexity in the number of dual variables and the size of $\MM(\vlam)$, they do not scale well to large problems. 
In our case $\MM(\vlam)$ is a $(12(\nrObsNodes)+1) \times (12(\nrObsNodes)+1)$ matrix  and for $\nrNodes = 5^3$, the 
size of $\MM(\vlam)$ is already larger than $10^3 \times 10^3$.}


{\bf Primal optimal solution via the dual.}  Here we demonstrate experimentally that one can recover a primal optimal solution $\vxxStar$ for the SLAM problem from the dual optimal solution $\vlam^\star$ of \eqref{eq:dualProblem} whenever strong duality holds.  

For each of the experimental trials described previously,
we also computed an estimate $\candidate$ for the SLAM solution directly from the optimal solution $\vlam^\star$ for the dual program \eqref{eq:dualProblem} using equation \eqref{eq:primalSolutionComputation}.   Figs.~\ref{fig:joint}(b1)-(b3) compare the value $f(\candidate)$ of this estimate against the value $\fMLstar$ of the solution 
obtained by solving the SLAM problem directly using the chordal initialization, and against the cost of the odometric guess.

We can see that for those experimental conditions in which strong duality holds (those experiments in Figs.~\ref{fig:joint}(a1)-(a3) for which the green and red bars are the same height), the estimate $\candidate$ in fact achieves the (certified) globally optimal cost $\fMLstar$ and is therefore a globally optimal solution for the SLAM problem, as guaranteed by Proposition~\ref{prop:primalOptimalViaDual}.  More interestingly, we also find that even in those cases when strong duality does \emph{not} hold (in which case Proposition~\ref{prop:primalOptimalViaDual} no longer guarantees that $\candidate$ is a primal optimal solution), the quality of the candidate $\candidate$  degrades gracefully (i.e.\ the gap $f(\candidate) - \fMLstar$ increases gradually) with increasing noise levels.  In particular, we find that $\candidate$ outperforms the odometric guess in all tested cases.

These experiments confirm that we can extract a primal optimal solution $\vxxStar$ for the SLAM problem \emph{directly} from the solution $\vlamStar$ of the convex dual problem whenever strong duality holds.  However, this approach is unfortunately not currently practical as a general-purpose SLAM technique, due to the high computational cost of solving large scale SDPs.

%
%
%
%
%

{\bf Effectiveness of \verificationTwo.}  In this section we show that \verificationTwo is a computationally
tractable verification approach, 
and it preserves the desirable properties of \verificationOne, i.e., it is able to discern optimal 
estimates from suboptimal ones. 
In contrast to \verificationOne, \verificationTwo 
does not quantify the sub-optimality gap, but can only give a binary answer: 
either it certifies that the estimate $\candidate$ is optimal (by producing a dual certificate $\vlamHat$), or it is inconclusive. 
Before moving to the real tests, 
we consider the \grid scenario discussed in the previous section.

We perform the following test: for each realization of the \grid scenario, we compute 
the optimal solution $\xopt$ (attaining $\fMLstar$) as in the previous tests, and a sub-optimal 
solution $\xsubopt$ (attaining $\fsubopt > \fopt$) by bootstrapping the Gauss-Newton method with a random initial guess. 
Then, we apply the verification technique to both $\xopt$ and $\xsubopt$ and see if they pass the 
optimality test. Using the results we can compute the \emph{precision} and \emph{recall} of our classification:
\bea
\text{precision} = \frac{|\accepted \xopt|}{|\accepted \xopt| + |\accepted \xsubopt|}
\;\;
\text{recall} = \frac{|\accepted \xopt|}{|\accepted \xopt| + |\rejected \xopt|}
\eea 

\normalsize
\noindent
where $|\accepted \xopt|$ denotes the number of tests in which an optimal 
solution $\xopt$ was accepted as optimal by \verificationTwo,
 $|\accepted \xsubopt|$ is the number of tests in which a suboptimal solution 
 was accepted as optimal, and $|\rejected \xopt|$ is the number of tests 
 in which  \verificationTwo was not able to certify the optimality of an optimal solution.

 We point out that \prettyref{prop:verificationTechniques}(\verificationTwo) guarantees that our certification approach always has precision equal to $1$; however, recall may be less than $1$, and will be for cases in which strong duality does not hold.   We plot precision/recall for different levels of translational and rotational noise, 
and for different sizes of the dataset, as shown in Figs.~\ref{fig:joint}(c1)-(c2)-(c3).  We observe that only for larger rotational noise Fig.~\ref{fig:joint}(c2) 
the recall decreases; these are exactly the cases in which the duality gap is nonzero.
  (For $\sigmaR = 0.2\mbox{rad}$, \verificationTwo 
 was not able to certify optimality in any case, which means that the precision becomes 
 undefined ($\frac{0}{0}$) and for this reason we do not show the corresponding data point in 
 Fig.~\ref{fig:joint}(c2).) Finally, Fig.~\ref{fig:joint}(d2) shows 
 the CPU time required by \verificationTwo. This is the time required to solve the linear system~\eqref{eq:verification2}, 
 and check if $\MM(\vlamHat) \succeq 0$. \verificationTwo is computationally cheap, as 
it does not require solving the SDP.

We conclude the experimental part of this paper by 
testing the performance of the verification technique \verificationTwo 
on large-scale SLAM datasets. We consider the same datasets as~\cite{Carlone15icra-init3D}:
the \sphere, \sphereHard, \torus and \grid are simulated datasets, while 
the \garage,  \cubicle and \rim are real datasets. 
For the results in this paper we substituted the covariances in the datasets 
of the scenarios \sphere, \sphereHard, \garage,  \cubicle and \rim with isotropic ones,
as required by the \PGO formulation~\eqref{eq:PGO}.

Table~\ref{tab:real} shows the results of the application of \verificationTwo to the
SLAM datasets. The first column shows the cost obtained by applying a \GN
method starting from the initialization~\cite{Carlone15icra-init3D} (rows ``Init.'' in the table) 
and from the odometric guess (rows ``Odom.''). These are the candidate solutions that we want to check, 
using \verificationTwo.
The columns 
 ``$d(\vlamHat)$'' and ``$\mu$'' show intermediate results of~\verificationTwo. 
In particular, $d(\vlamHat)$ is the same one described in \prettyref{prop:verificationTechniques}(\verificationTwo), 
while $\mu$ is the smallest eigenvalue of $\MM(\vlamHat)$. 
Recall that \verificationTwo certifies optimality when $\fML(\hat{\vxx}) = d(\vlamHat)$ 
and $\mu \geq 0$ (if the smallest eigenvalue is non-negative, then $\MM(\vlamHat) \succeq 0$).
We specify a tolerance in these tests, since the \GN estimate $\hat{\vxx}$ will not attain  the  optimal solution ${\vxx}^\star$ exactly.  Consequently, 
the solution $\vlamHat$ of~\eqref{eq:verification2} is not 
 exact, so we consider that $\fML(\hat{\vxx}) = d(\vlamHat)$ if 
 $|\fML(\hat{\vxx}) - d(\vlamHat)| / \fML(\hat{\vxx}) < 20\%$. 
 Similarly, we add some tolerance to the condition 
 $\mu \!\geq\! 0$, and accept $\mu \!\!\geq\!\! -1$. Recall that $\mu$ is expected to be slightly negative, as the objective in~\eqref{eq:dualProblem} 
 tends to push $\MM(\vlambda)$ towards the boundary of the positive definite cone, and the SDP is solved numerically.


\renewcommand{\e}[1]{\!\cdot\!10^{#1}\!}

\newcommand{\good}[1]{{\color{blue}#1}}
\newcommand{\no}{$-$}
\newcommand{\myfont}[1]{{\small #1}}
\newcommand{\brac}{\bigg\}}
\newcommand{\fang}{f_{\mbox{ang}}}
\newcommand{\fchord}{f_{\mbox{chord}}}

\newcommand{\showMuFull}[1]{} 

\begin{center}
\begin{table}[t]
\renewcommand*\arraystretch{1.1}
\begin{centering}
{\scriptsize 
\begin{tabular}{| m{0.9cm} | m{0.5cm} | m{1.3cm} | m{1.1cm} | m{1.35cm}  | m{0.45cm} | m{0.1cm} |}
\hline
&  & $\fML(\hat\vxx)$ & $d(\vlamHat)$ & $\mu$ \showMuFull{\multicolumn{2}{c|}{$\mu^{-}$}}  & Time & \verificationTwo \\
\hline
\hline
\!\!\sphere & Init. & $5.7595\e{2}$  & $5.75\e{2}$ & $-1.80\e{-4}$ \showMuFull{& ($-1.81\e{-4}$)} & $609$ & \green{\checkmark} \\ 
$\substack{ n=2500 \\ m=4949}$ & Odom. & \red{$5.8019\e{2}$}  & \red{$4.38\e{2}$} & $-1.45\e{-4}$ \showMuFull{& ($-1.45\e{-4}$)} & $597$ & \red{X} \\ 
\hline 
\!\!\sphereHard & Init. & $1.2485\e{6}$  & $1.25\e{6}$ & $-9.64\e{-3}$ \showMuFull{& ($-9.64\e{-3}$)} & $332$ & \green{\checkmark} \\ 
$\substack{ n=2200 \\ m=8647}$ & Odom. & \red{$3.0413\e{6}$}  & $3.04\e{6}$ & \red{$-1.01\e{2}$} \showMuFull{& (\red{$-5.82\e{3}$})} & $2.5$ & \red{X} \\ 
\hline 
\!\!\torus & Init. & $1.2114\e{4}$  & $1.21\e{4}$ & $-7.85\e{-2}$ \showMuFull{& ($-7.85\e{-2}$)} & $39$ & \green{\checkmark} \\ 
$\substack{ n=5000 \\ m=9048}$ & Odom. & \red{$2.7666\e{4}$}  & $2.76\e{4}$ & \red{$-1.04\e{2}$} \showMuFull{& (\red{$-1.37\e{2}$})} & $3.7$ & \red{X} \\ 
\hline 
\!\!\grid & Init. & $4.216\e{4}$  & $4.22\e{4}$ & $-1.24\e{-2}$ \showMuFull{& (--)} & $1646$ & \green{\checkmark} \\ 
$\substack{ n=8000 \\ m=22236}$ & Odom. & \red{$2.7465\e{5}$}  & $2.74\e{5}$ & \red{$-9.98\e{1}$} \showMuFull{& (--)} & $68.2$ & \red{X} \\ 
\hline 
\!\!\garage & Init. & $\!\!6.2994\e{-1}$  & $6.11\e{-1}$ & $-6.78\e{-2}$ \showMuFull{& ($-6.78\e{-2}$)} & $12.3$ & \green{\checkmark} \\ 
$\substack{ n=1661 \\ m=6275}$ & Odom. & $\!\!6.2997\e{-1}$  & \red{$3.53\e{-1}$} & $-6.75\e{-2}$ \showMuFull{& ($-6.75\e{-2}$)} & $11.2$ & \red{X} \\ 
\hline 
\!\!\cubicle & Init. & $\!\!6.2481\!\e{2}\!$  & $6.25\e{2}$ & $-4.76\e{-1}$ \showMuFull{& (--)} & $16.8$ & \green{\checkmark} \\ 
$\substack{ n=5750 \\ m=16869}$ & Odom. & $6.2484\e{2}$  & $6.16\e{2}$ & $-4.75\e{-1}$ \showMuFull{& (--)} & $15.5$ & \green{\checkmark} \\ 
\hline 
\!\!\rim & Init. & $1.235\e{4}$  & $1.23\e{4}$ & \red{$-9.77\e{1}$} \showMuFull{& (--)} & $7.7$ & \red{X} \\ 
$\substack{ n=10195 \\ m=29743}$ & Odom. & \red{$1.6985\e{4}$}  & \red{$-2.8\e{4}$} & \red{$-9.38\e{1}$} \showMuFull{& (--)} & $7.1$ & \red{X} \\ 
\hline 
\end{tabular}
}
\caption{Verification technique \verificationTwo on large-scale SLAM datasets.
\label{tab:real}}
\end{centering}
\vspace{-0.8cm}
\end{table}
\end{center}
 
\vspace{-\baselineskip}

Let us start our analysis from the \sphere dataset. When using the initialization (``Init.'' row), 
 the \GN method attains an objective $5.7595\e{2}$.  
 In this case, the two conditions for \verificationTwo 
 are satisfied and we can certify the optimality of the resulting estimate (green check-mark in the rightmost column). 
  When using the odometric initialization the resulting cost is larger (red entry in the column $\fML(\hat\vxx)$), 
  hence the estimate is suboptimal. \verificationTwo is able to identify the suboptimality, since $d(\vlamHat)$ 
  becomes much smaller than $\fML(\hat\vxx)$. Hence \verificationTwo correctly decides not to certify the optimality of 
  the odometric estimate (red ``\red{X}'' in the last column). 
  Similar considerations hold for the second scenario, the challenging \sphereHard: in this case 
  the initialized estimate (shown in Fig.~\ref{fig:maps}, top left) is accepted as optimal. The odometric estimate, 
  instead, is trapped in a local minimum (Fig.~\ref{fig:maps}, bottom left), and the optimality test \verificationTwo 
  correctly rejects the estimate since it leads to a very negative $\mu$.  
  Similar considerations hold for the \torus, \grid, and the \cubicle datasets: our technique is able to 
  discern optimal solutions from suboptimal estimates in all cases.
  
  For the \garage dataset, the initialized estimate is classified as optimal, while the 
  odometric estimate, which has a similar cost ($6.2997\e{-1}$ {\tt vs} $6.2994\e{-1}\,$), is 
  rejected as suboptimal. We observed the corresponding trajectory estimates 
  (see~Fig.\ref{fig:garage}) and, while they are 
  are both visually correct and have very similar costs, they do not overlap. 
  This may indicate the presence of regions in the cost functions that are nearly flat, 
  i.e., for which different estimates can have similar cost. Allowing for extra \GN iterations, 
  the odometric estimate converges to the same cost as the initialized one, and our technique 
  is able to certify its optimality. Empirically, we observed that estimates that are suboptimal 
  because they need extra iterations to converge 
  tend to fail the check on $d(\vlamHat)$ 
  (compare with \sphere and \garage), while estimates that converge to a 
  wrong minimum tend to fail the check on $\mu$ (compare with \sphereHard, \torus, and \grid).
  
  For the \rim dataset, both estimates (``init'' and ``odom'') are rejected by \verificationTwo. 
  While the initialized estimate is visually correct, currently, we cannot conclude anything as 
  it might be that there exists a better estimate (i.e., \GN failed), or our technique failed 
  because the duality gap was nonzero.  
  
  Finally, in the column ``Time'', we report the CPU time (in seconds) required to 
  perform the second verification technique. Most of the time here is spent computing 
  the smallest eigenvalue $\mu$ of $\MM(\vlamHat)$, which was obtained using  
   Matlab's ${\tt eigs}$, specifying $-100$ as guess for the eigenvalue. 
  The CPU time depends on the size of the problem, but 
  also depends on the distance between our guess ($-100$) and the closest eigenvalue.
  We leave a more thorough investigation of these computational aspects for future work.

%
%


\section{Conclusion}
\label{sec:conclusion}

We show that Lagrangian duality is an effective tool to assess the quality 
of a given SLAM estimate. We propose two techniques to judge if an estimate is 
globally optimal (i.e., it is the \ML estimate), and we show that when the duality gap is zero
one can compute an optimal SLAM solution from the dual problem.
The performance of our verification techniques is extensively tested in real and simulated 
datasets, including large-scale benchmarks.
Many theoretical questions remain open, e.g., under which conditions the duality gap is zero? 
is it possible to derive bounds on the duality gap? 
We also leave as future work more practical problems, such as the one of exploring more 
efficient solvers for large SDPs, possibly exploiting problem structure.

\bibliographystyle{IEEEtran}
\bibliography{refs}

\newpage

\appendix

\subsection{Extended Proof of~\prettyref{lem:xstarInNullSpace}}
\label{sec:appendix}


\section{Extended Proof of~\prettyref{lem:xstarInNullSpace}}
\label{sec:appendix}

Here we give an alternative proof of~\prettyref{lem:xstarInNullSpace}. 
For the sake of readability, we restate the Lemma before proving it. 

\vspace{0.1cm}
\begin{lemma}[Primal optimal solution and zero duality gap]
If the duality gap is zero ($d^\star = f^\star$), then any 
primal optimal solution $[\vxx^\star \; 1]$ of~\eqref{eq:primal} is in the null space of the 
matrix $\MM(\vlambda^\star)$, where $\vlambda^\star$ is the solution of the dual problem~\eqref{eq:dualProblem}.
\end{lemma}
\vspace{0.1cm}

\begin{IEEEproof} 
Assume that the duality gap between the primal problem~\eqref{eq:primal} 
and the dual problem~\eqref{eq:dualProblem} is zero:
\beq
\label{eq:proof10}
d^\star = f^\star 
\Leftrightarrow 
\sum_{ \substack{i=\obsNodes \\ u=1,2,3} } \lamParaStar_{iuu} + \lamyStar =  \normsq{ \MAanc \vxx^\star - \vb y^\star}{} 
\eeq
Now, since $[\vxxStar \; \yStar]$ is a solution of the primal problem, it must also be feasible, hence 
the following equality holds:
\bea
\label{eq:proof11}
&\sumObsNodes \bigg[
\displaystyle \sum_{ \substack{ u=1,2,3} } \lamParaStar_{iuu} (1 - (\vxxStar)\tran \MEanc_{iuv} (\vxxStar))  \\ 
&+ \!\!\!\! \displaystyle \sum_{ \substack{ u,v=1,2,3 \\ u \neq v} } \lamPerpStar_{iuv} (-(\vxxStar)\tran \MEanc_{iuv} (\vxxStar) )
\bigg]
+ \lamyStar (1 - (\yStar)^2)= 0 \nonumber
\eea
Therefore, we can subtract the left-hand side of~\eqref{eq:proof11} to the left-hand side of~\eqref{eq:proof10} without altering the result:
\bea
\label{eq:proof12}
- \sumObsNodes \bigg[
\displaystyle \sum_{ \substack{ u=1,2,3} } \lamParaStar_{iuu} (1 - (\vxxStar)\tran \MEanc_{iuv} (\vxxStar)) \nonumber \\ 
- \!\!\!\! \displaystyle \sum_{ \substack{ u,v=1,2,3 \\ u \neq v} } \lamPerpStar_{iuv} (-(\vxxStar)\tran \MEanc_{iuv} (\vxxStar) ) \bigg]
- \lamyStar (1 - (\yStar)^2)  
\nonumber \\
+ \sum_{ \substack{i=1,\ldots,n \\ u=1,2,3} } \lamParaStar_{iuu} + \lamyStar =  \normsq{ \MAanc \vxx^\star - \vb y^\star}{} 
\eea
Noting that some terms  in $\lamParaStar_{iuu}$ and $\lamyStar$ cancel out, we get:
\bea
\label{eq:proof13}
 \sumObsNodes & \!\!\!\! \bigg[
\displaystyle \sum_{ \substack{ u=1,2,3} } \!\! \lamParaStar_{iuu} (\vxxStar)\tran \MEanc_{iuv} (\vxxStar) +
\!\!\!\! \displaystyle \sum_{ \substack{ u,v=1,2,3 \\ u \neq v} } \!\! \lamPerpStar_{iuv} (\vxxStar)\tran \MEanc_{iuv} (\vxxStar)  \bigg] \nonumber \\ 
& + \lamyStar (\yStar)^2  =  \normsq{ \MAanc \vxx^\star - \vb y^\star}{}  \nonumber
\eea
Reorganizing the terms as done in~\eqref{eq:quadraticTerms}, we obtain:
\bea
\label{eq:proof13}
\vect{\vxxStar \\ \yStar}\tran 
\matTwo{\MH(\vlambda^\star)
 & -\MAanc\tran \vb  \\
- \vb\tran \MAanc &   \vb\tran \vb - \lamyStar} 
 \vect{\vxxStar \\ \yStar} = 0
\eea
which implies that $[\vxxStar \; \yStar]$ is the null space of $\MM(\vlamStar)$, proving the claim.
 \end{IEEEproof}

\newpage

\subsection{Estimated Trajectories for the Datasets in Table~\ref{tab:real}}
\label{sec:visualization} 


\definecolor{dgreen}{rgb}{0,0.5,0}

\renewcommand{\widthCol}{3cm}

\vspace{0.5cm}

\begin{figure}[h]
\vspace{-0.5cm}
\begin{minipage}{8.5cm}
\begin{tabular}{ccc}%
& \color{blue}{\GN from initialization} & \hspace{0.5cm}  \color{red}{\GN from odometry}
\vspace{-0.1cm}
\\
\begin{sideways}{$\!\!\!\!${\sphere}}\end{sideways} &  
\begin{minipage}{\widthCol}%
\centering%
\includegraphics[scale=0.3, trim=5cm 1cm 0cm 2cm,clip]{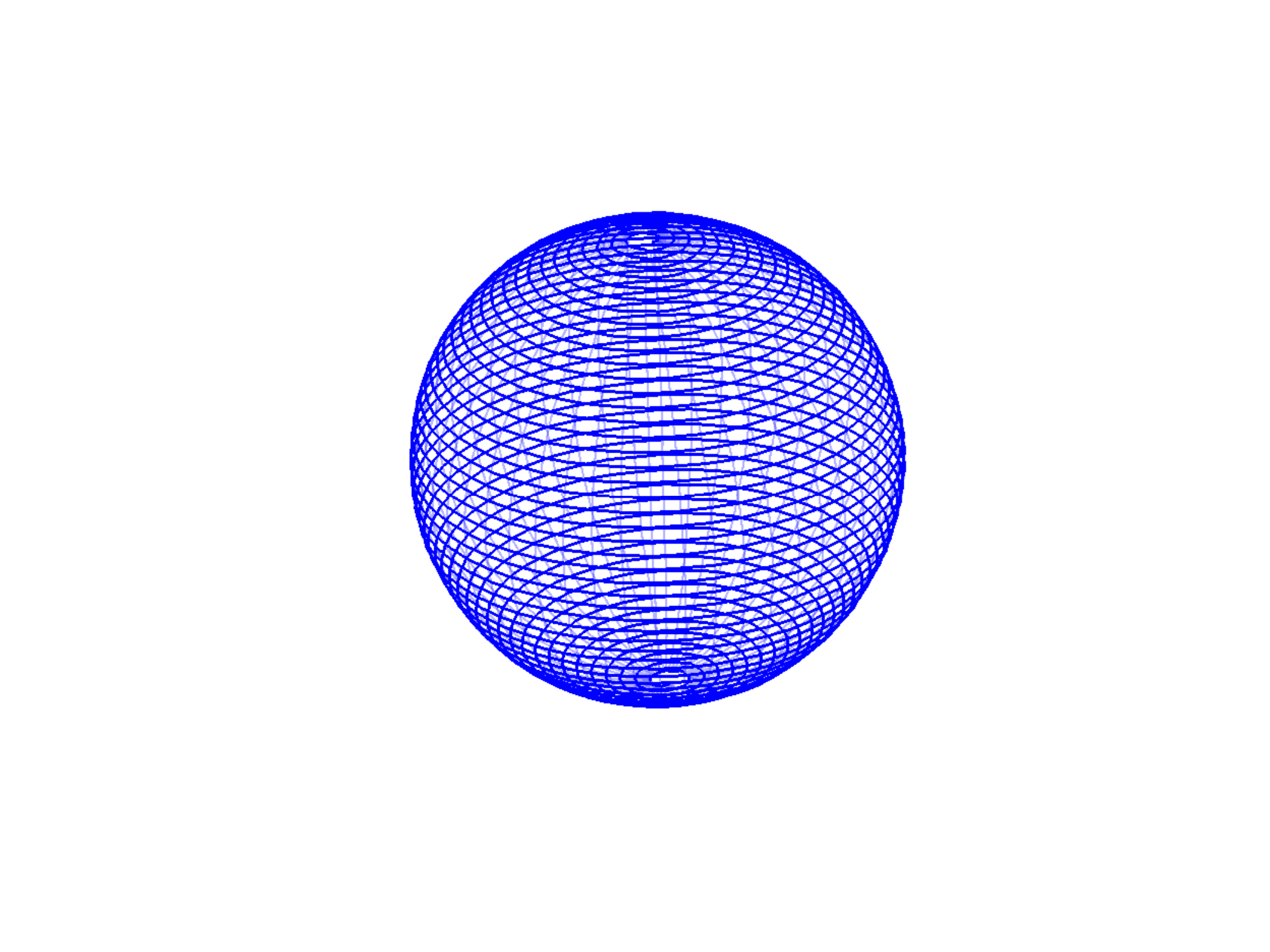} 
\end{minipage}
& 
\begin{minipage}{\widthCol}%
\centering%
\includegraphics[scale=0.3, trim=4cm 1cm 1cm 2cm,clip]{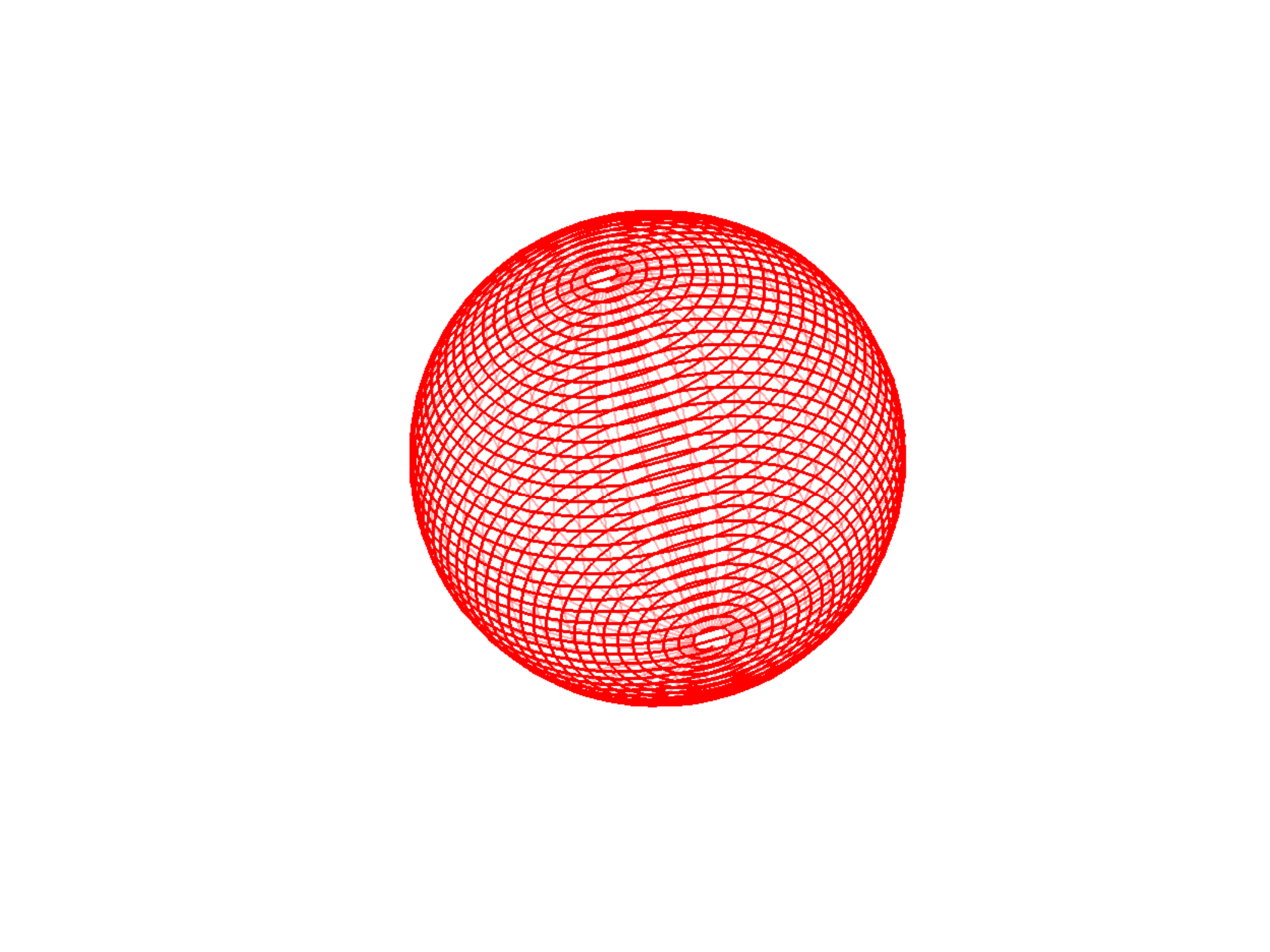}  
\end{minipage}
\vspace{-0.8cm}
\\
\begin{sideways}{$\!\!\!\!\!\!\!\!\!\!\!${\sphereHard}}\end{sideways} & 
\begin{minipage}{\widthCol}%
\centering%
\includegraphics[scale=0.32, trim=5cm 1cm 0cm 1cm,clip]{figuresMatlab-refinedPoses-gtsam-sphere_bignoise_vertex3-chordal-eps-converted-to.pdf} 
\end{minipage}
& 
\begin{minipage}{\widthCol}%
\centering%
\includegraphics[scale=0.25, trim=4cm 0cm 0cm 1cm,clip]{figuresMatlab-refinedPoses-odometry-sphere_bignoise_vertex3-X-eps-converted-to.pdf}  
\end{minipage}
\vspace{-0.8cm}
\\
\begin{sideways}{$\!\!\!\!\!\!\!\!\!\!\!${\torus}}\end{sideways} & 
\begin{minipage}{\widthCol}%
\centering%
\includegraphics[scale=0.27, trim=3cm 2cm 0cm 1cm,clip]{figuresMatlab-refinedPoses-gtsam-torus3D-chordal-eps-converted-to.pdf}  
\end{minipage}
& 
\begin{minipage}{\widthCol}%
\centering%
\includegraphics[scale=0.32, trim=4cm 2cm 2cm 1cm,clip]{figuresMatlab-refinedPoses-odometry-torus3D-X-eps-converted-to.pdf}
\end{minipage}
\vspace{-0.8cm}
\end{tabular}%
\vspace{-0.3cm} 
\end{minipage}%
\end{figure}

\begin{figure}[h!]
\begin{minipage}{8.5cm}
\begin{tabular}{ccc}%
\begin{sideways}{{\grid}}\end{sideways} & 
\begin{minipage}{\widthCol}%
\vspace{-0.5cm}
\centering%
\includegraphics[scale=0.24, trim=5cm 1cm 0cm 2cm,clip]{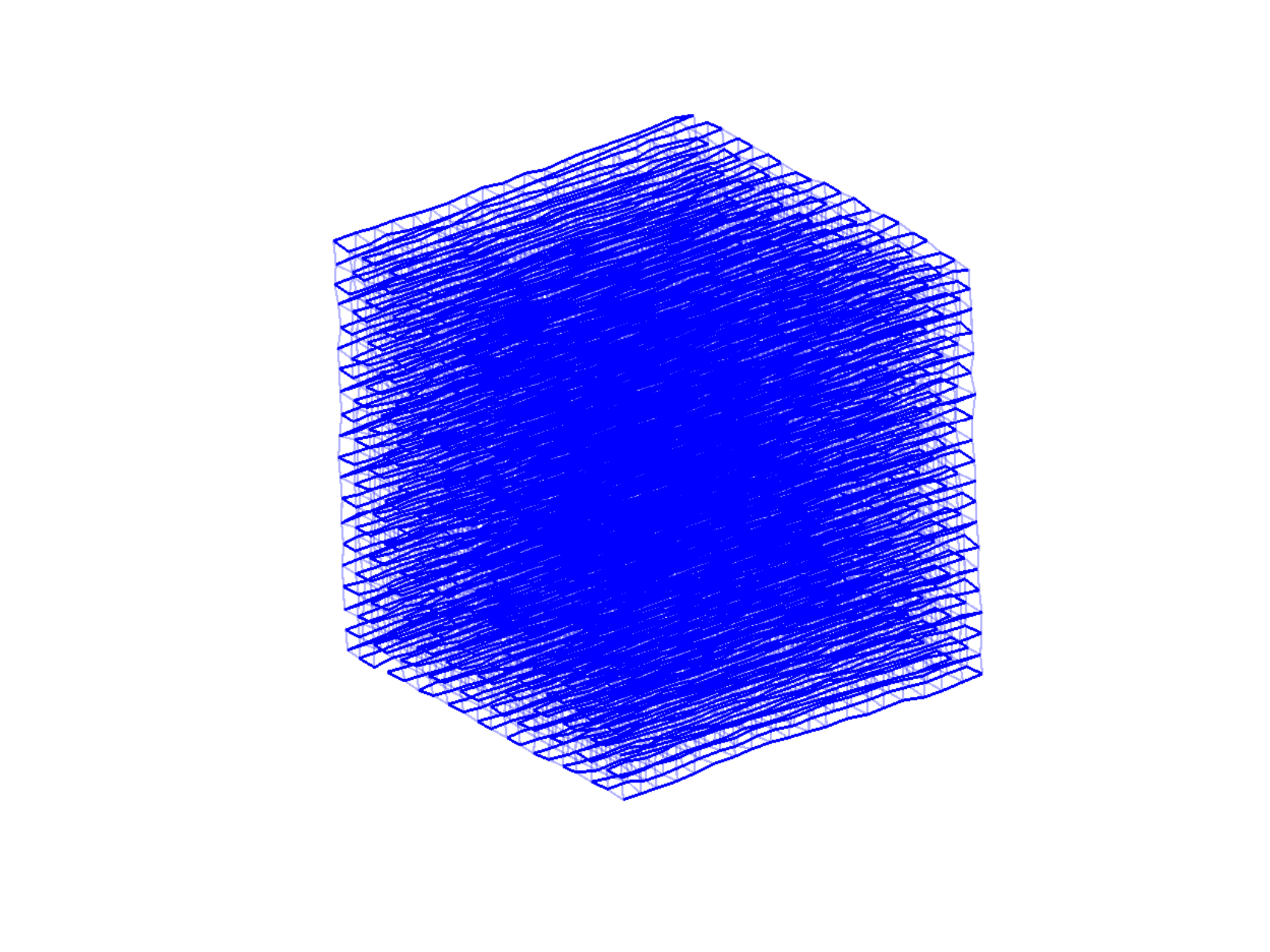} 
\end{minipage}
& 
\begin{minipage}{\widthCol}%
\centering%
\includegraphics[scale=0.35, trim=5cm 0cm 0cm 2cm,clip]{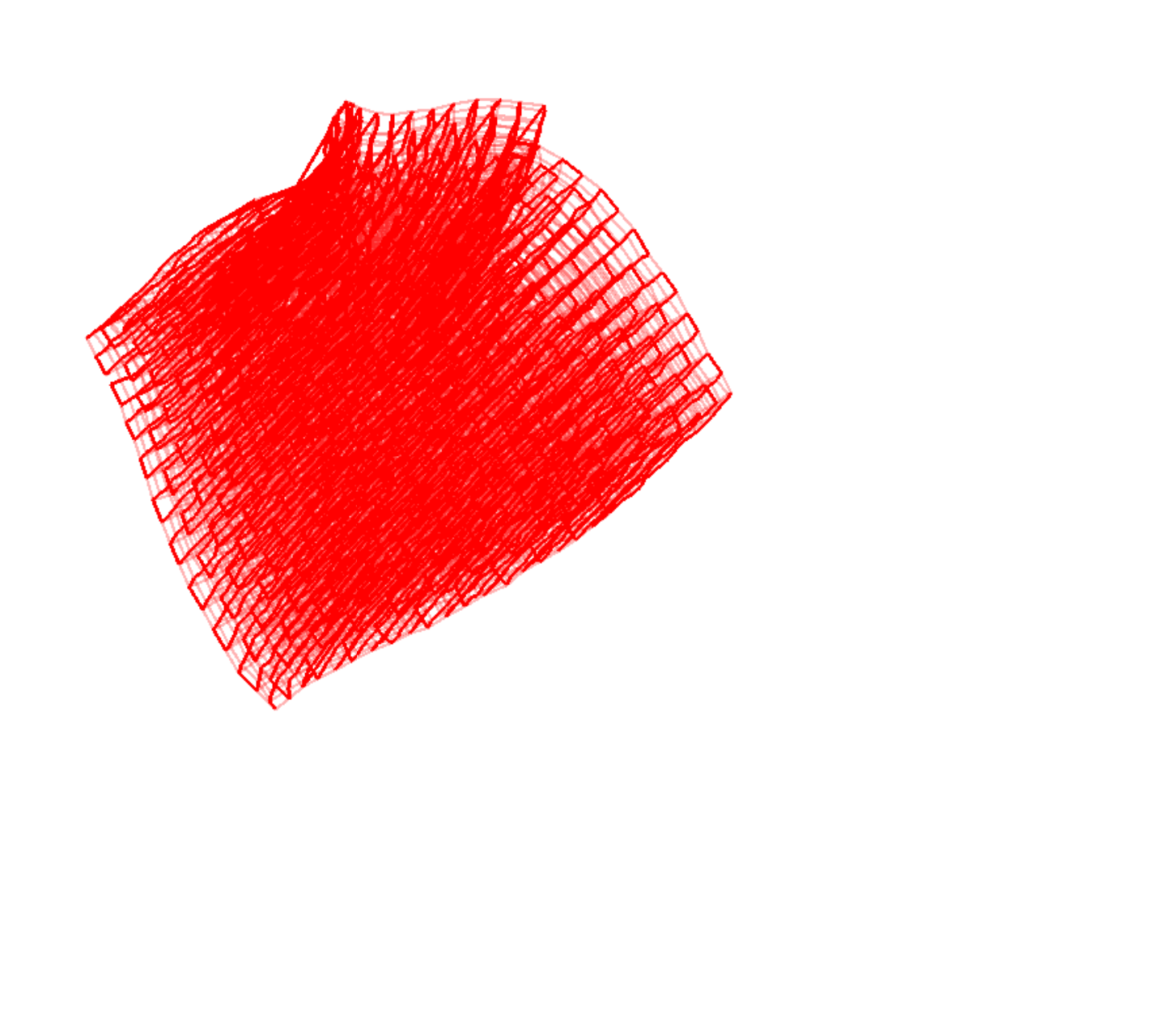}  
\end{minipage}
\vspace{-1.5cm}
\\
\begin{sideways}{$\!\!\!\!\!\!\!\!\!\!\!${\garage}}\end{sideways} & 
\begin{minipage}{\widthCol}%
\centering%
\includegraphics[scale=0.25, trim=6cm 3cm 0cm 2cm,clip]{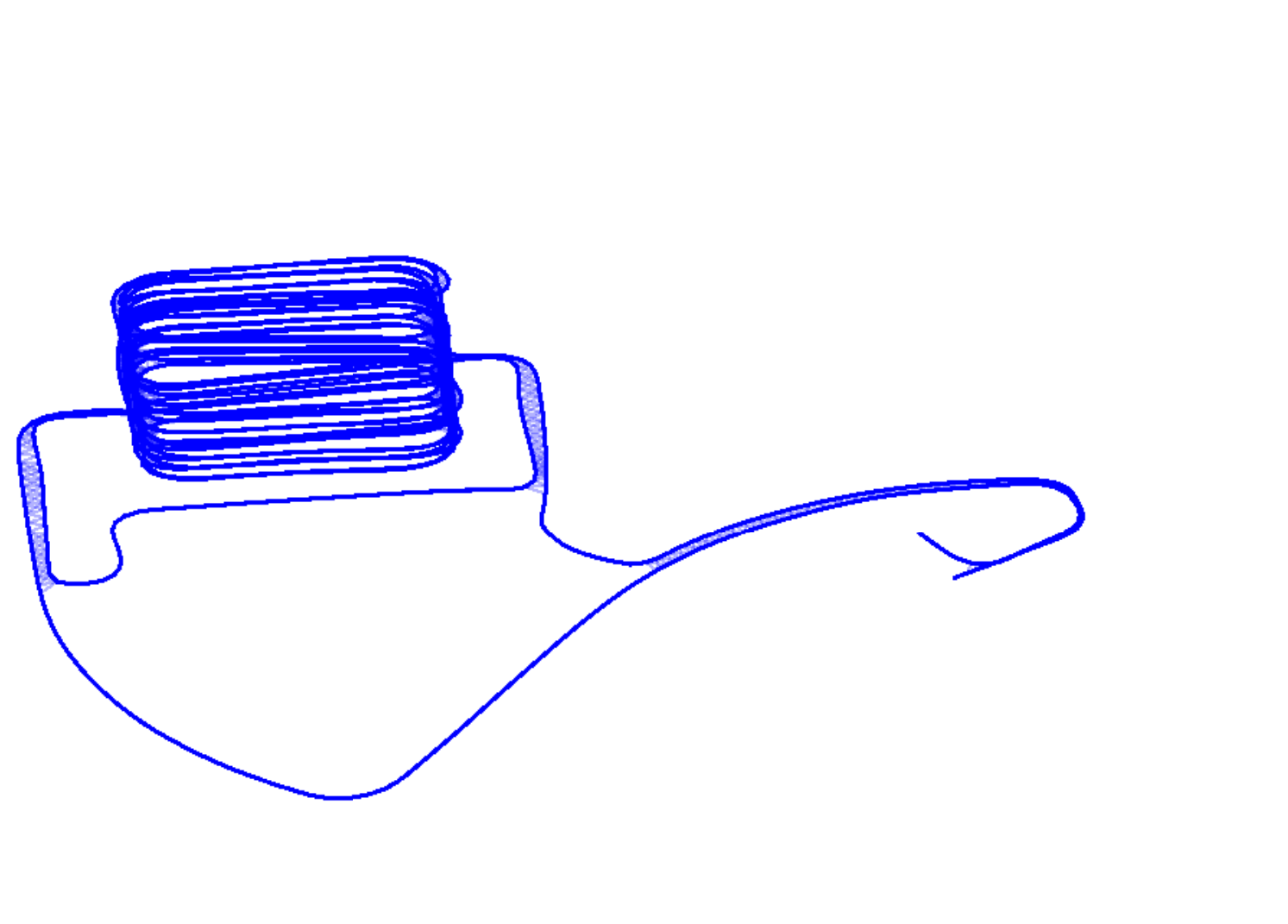} 
\end{minipage}
& 
\begin{minipage}{\widthCol}%
\centering%
\includegraphics[scale=0.27, trim=5cm 2cm 0cm 2cm,clip]{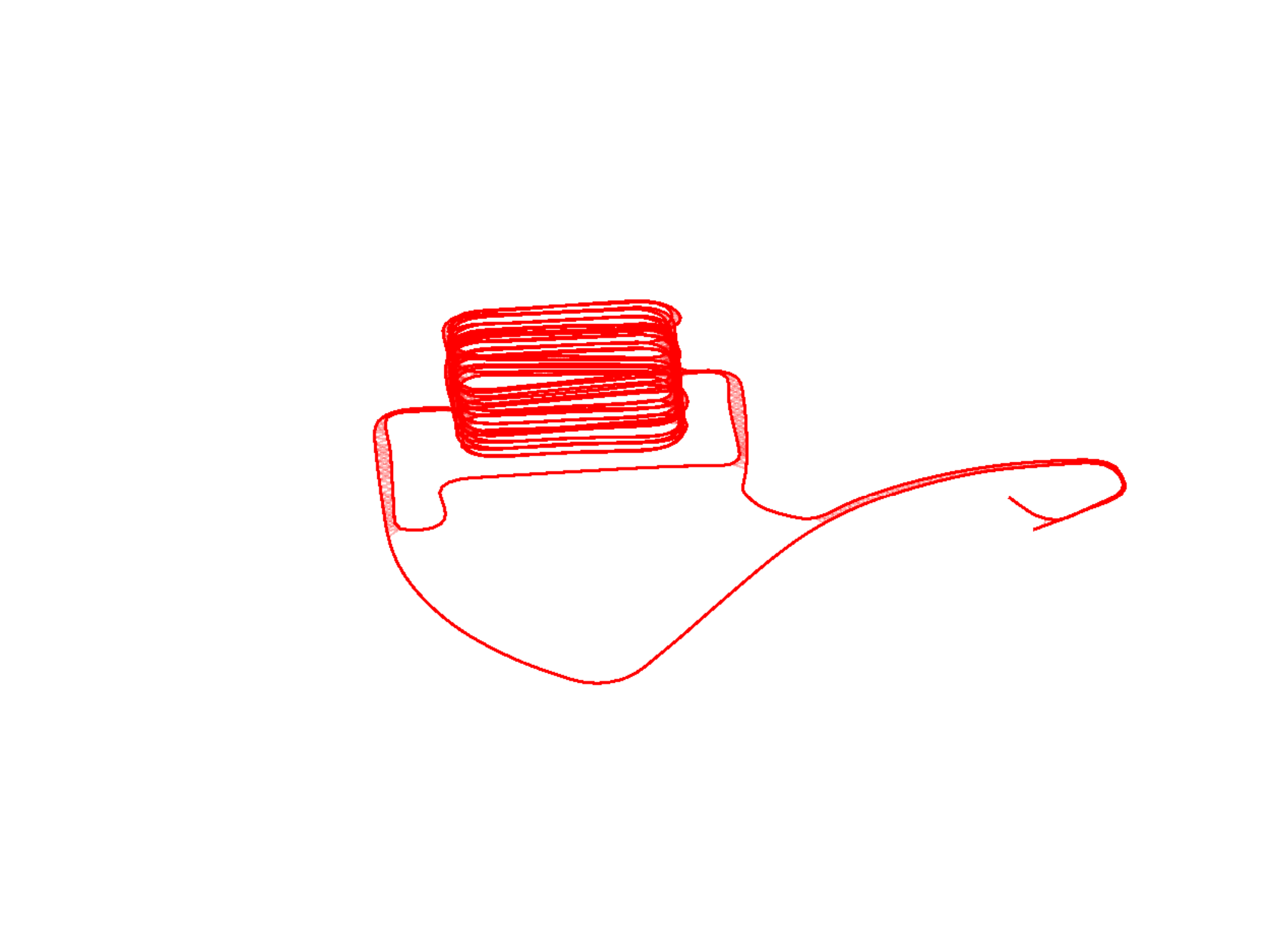}  
\end{minipage}
\vspace{-0.8cm}
\\
\\
\begin{sideways}{$\!\!\!\!\!\!\!\!\!\!\!${\cubicle}}\end{sideways} & 
\begin{minipage}{\widthCol}%
\centering%
\includegraphics[scale=0.25, trim=4.2cm 3cm 0cm 3cm,clip]{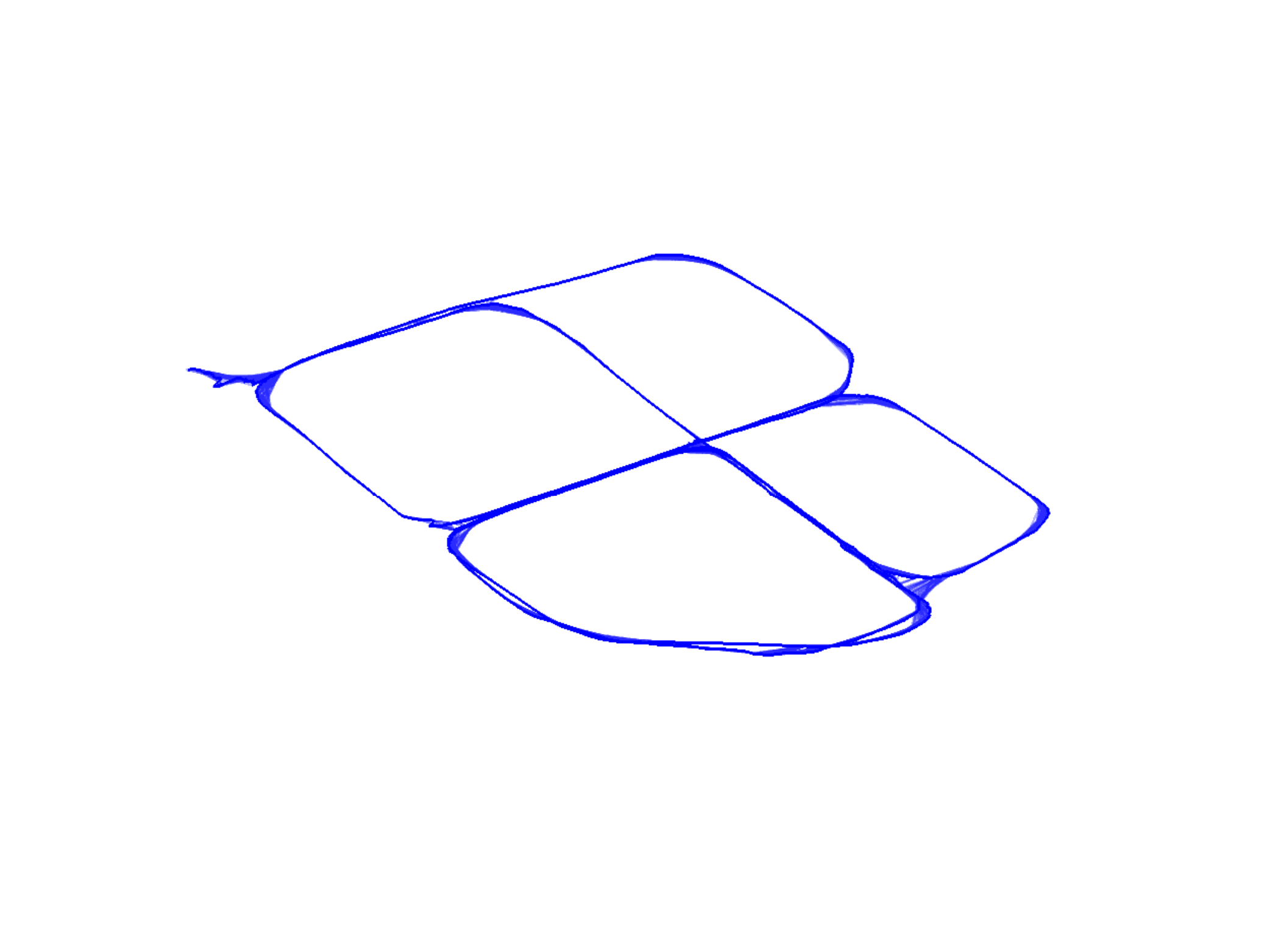} 
\end{minipage}
& 
\begin{minipage}{\widthCol}%
\centering%
\includegraphics[scale=0.3, trim=4cm 2cm 0cm 3cm,clip]{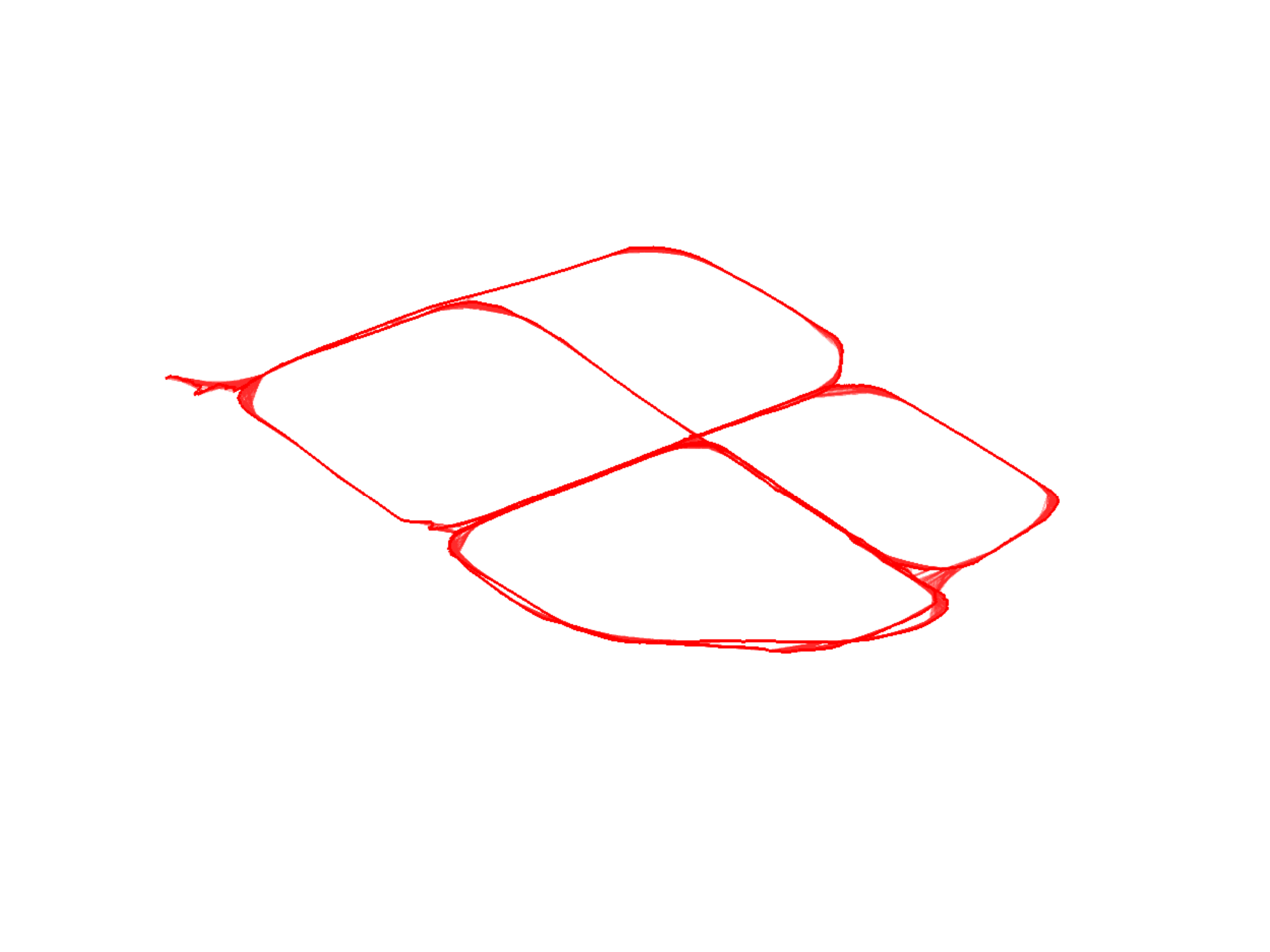}  
\end{minipage}
\vspace{-0.8cm}
\\
\\
\begin{sideways}{$\!${\rim}}\end{sideways} &
\begin{minipage}{\widthCol}%
\centering%
\includegraphics[scale=0.23, trim=3cm 2cm 0cm 2cm,clip]{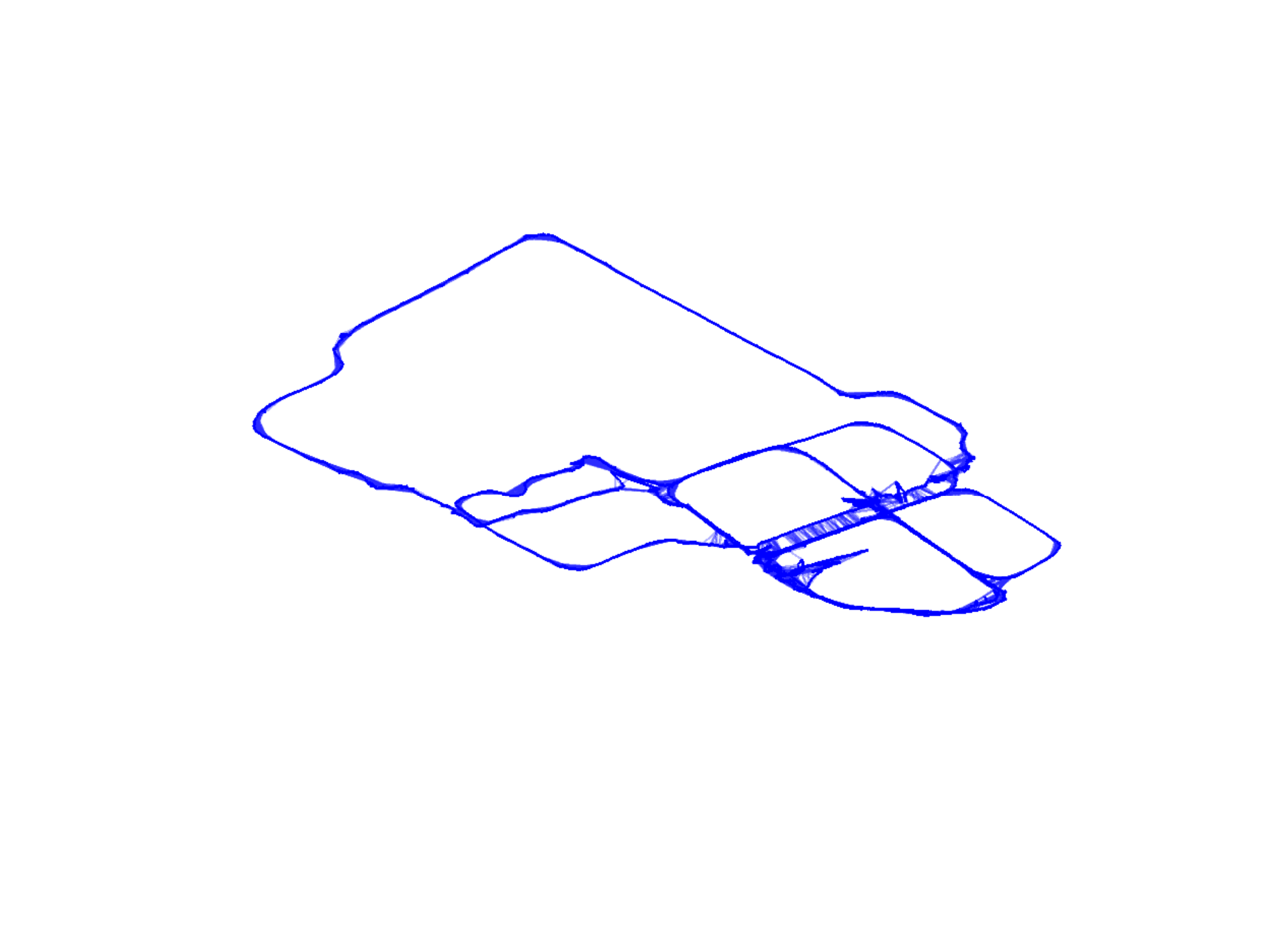} 
\end{minipage}
& 
\begin{minipage}{\widthCol}%
\centering%
\includegraphics[scale=0.32, trim=4cm 2cm 2cm 3cm,clip]{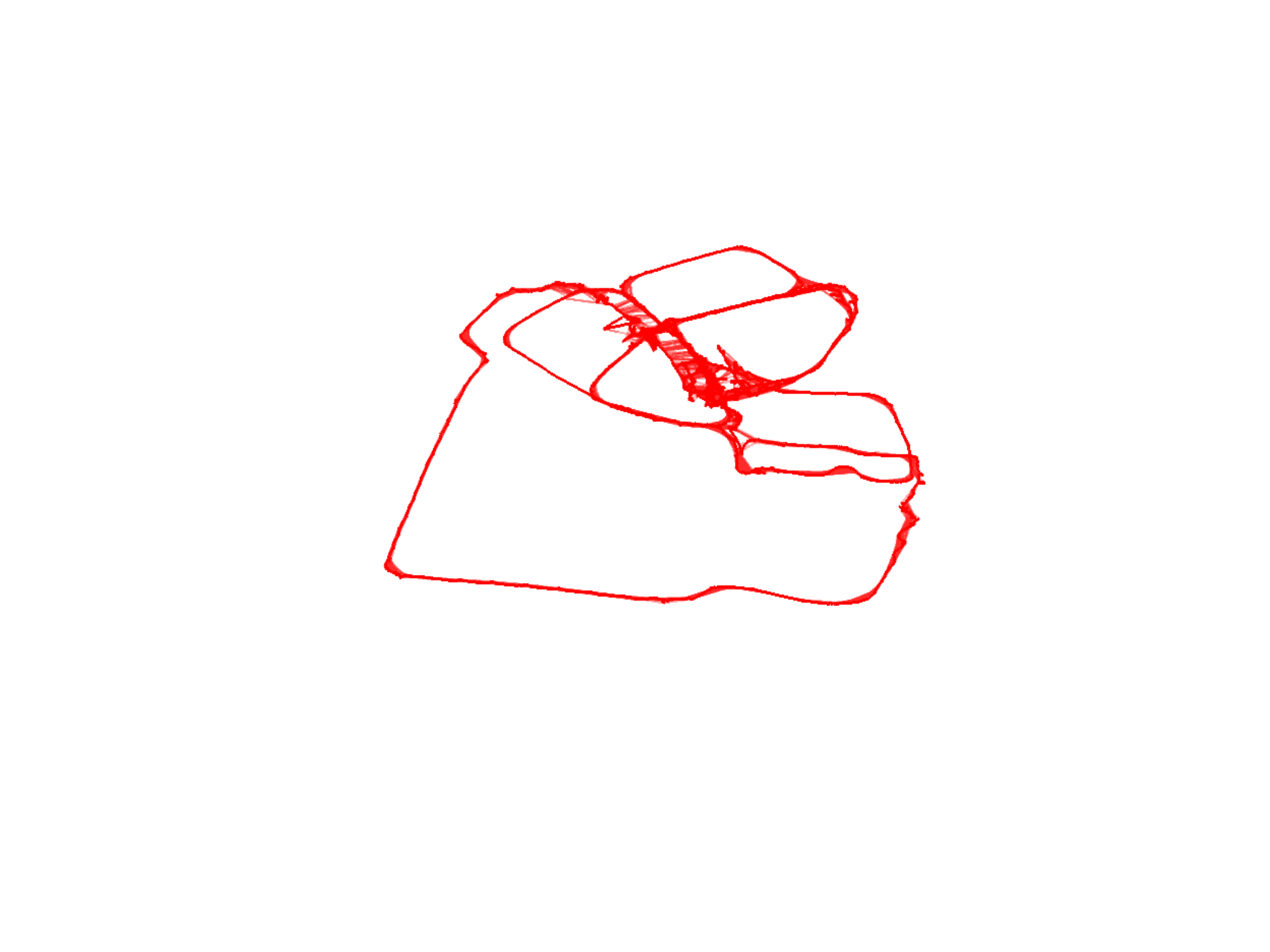}
\end{minipage}
\vspace{-0.8cm}
\end{tabular}%
\end{minipage}%
\caption{Blue: Estimated trajectory obtained by refining the \emph{chordal} initialization of~\cite{Carlone15icra-init3D} 
 with 10 Gauss-Newton iterations. Red: Estimated trajectory obtained by refining the odometric guess 
 with 10 Gauss-Newton iterations.}
\end{figure}

\newpage

\begin{figure}[h!]
\begin{minipage}{8.5cm}
\begin{tabular}{cc}%
\begin{sideways}{\hspace{-5cm}{\garage: zoomed-in view and comparison}}\end{sideways} & 
\begin{minipage}{8.5cm}%
\vspace{0.5cm}
\centering%
\includegraphics[scale=0.4, trim=5cm 1cm 0cm 2cm,clip]{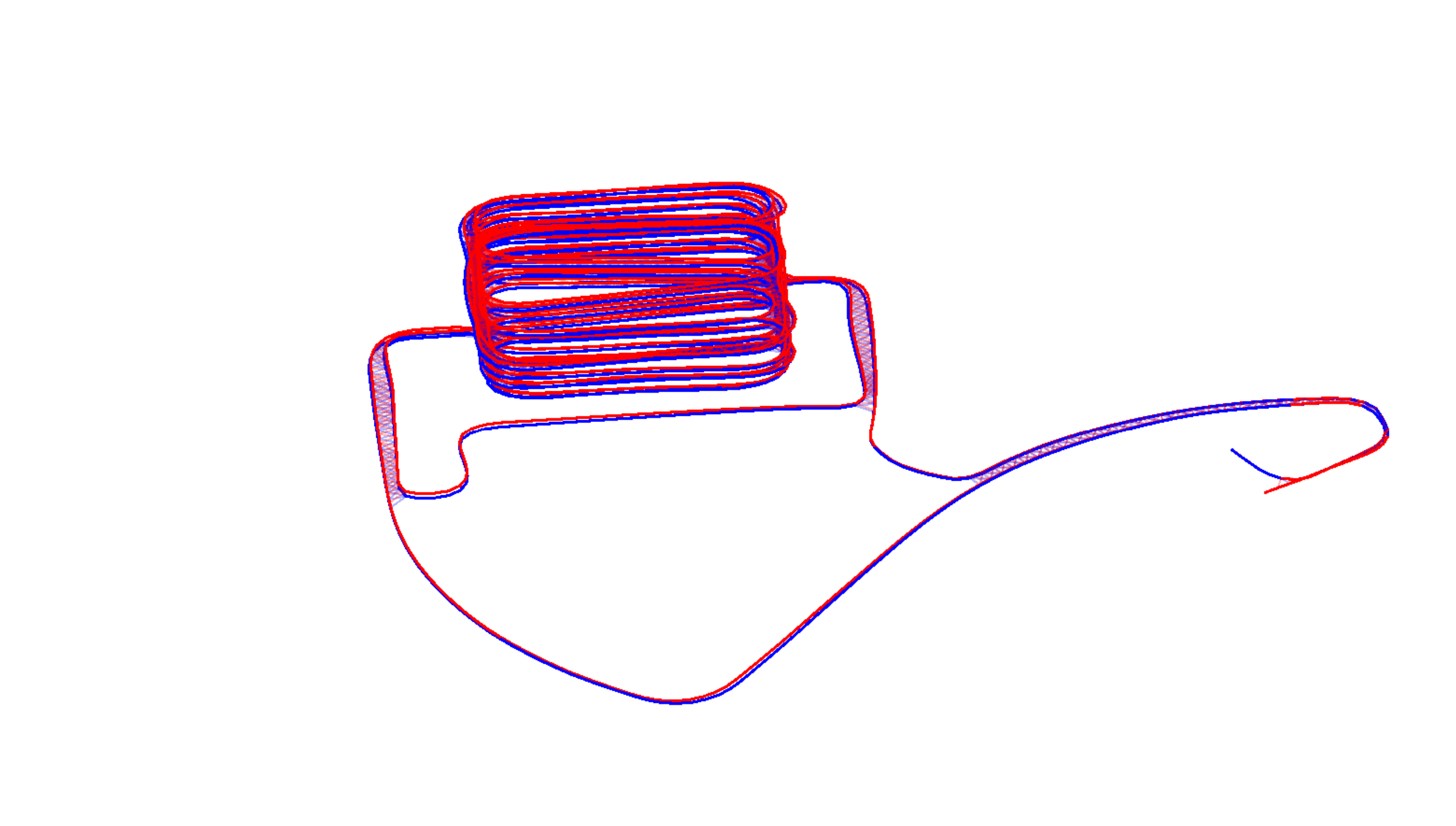} 
\end{minipage}
\\
& \begin{minipage}{8.5cm}%
\hspace{-1cm}
\centering%
\includegraphics[scale=0.25, trim=3cm 1cm 1cm 0cm,clip]{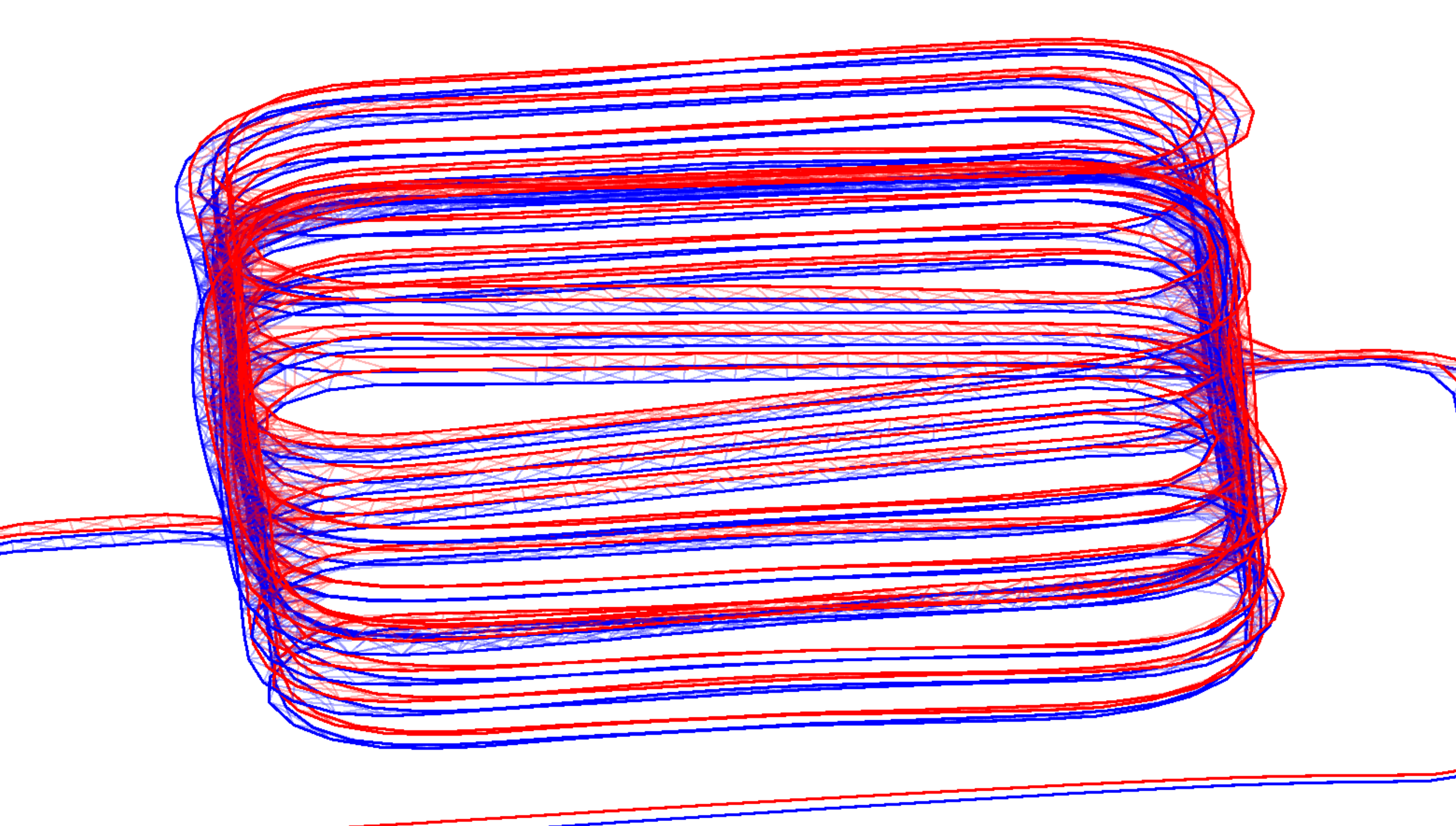}  
\end{minipage}
\end{tabular}%
\end{minipage}%
\caption{\garage dataset: while the estimates obtained from \GN with initialization (blue) and 
\GN from odometry (red) attain a very similar cost ($6.2994\e{-1}$ versus $6.2997\e{-1}$), 
the corresponding trajectories are still different, which indicates that \GN from odometry 
did not completely converge after 10 iterations.\label{fig:garage}}
\end{figure}

\vspace{1cm}

\end{document}